\documentclass[11pt]{article}

\usepackage[final]{acl}

\usepackage{times}
\usepackage{latexsym}
\usepackage{amsmath}
\usepackage{amssymb}
\usepackage{booktabs}
\usepackage{colortbl}
\usepackage{xcolor}
\usepackage{multirow}
\usepackage{makecell}
\usepackage{tabularx}
\usepackage{bbm}
\usepackage{pifont}
\usepackage[T1]{fontenc}
\usepackage[utf8]{inputenc}

\usepackage{microtype}

\usepackage{inconsolata}

\usepackage{graphicx}

\title{MoF: Preference-Aware Mixture Modeling for Black-Box LLM Personalization}

\author{Hun Park\\
  Department of Artificial Intelligence \\
  Korea University \\
  Seoul, Republic of Korea \\
  \texttt{hunbag98@korea.ac.kr} \\
  }

\begin{document}
\maketitle
\begin{abstract}
  Proprietary Large Language Models (LLMs) have demonstrated remarkable capabilities across a wide range of tasks, yet aligning their outputs with diverse user preferences remains challenging. 
  Existing personalization approaches for black-box LLMs often rely on user-specific scoring heads, causing the number of personalized parameters to grow linearly with the number of users and requiring additional adaptation for unseen users.
  To address these limitations, we propose \textbf{Mixture-of-Facets (MoF)}\footnote{Code is publicly available at \url{https://github.com/hoonpak/Mixture-of-Facets}.}, a scalable personalization framework for black-box LLMs that models user preferences as compositions of shared latent preference facets rather than dedicated user-specific parameters. 
  MoF performs personalization through history-conditioned routing over shared facet heads, enabling personalization for users unseen during training without additional parameter updates.
  Across diverse personalization tasks, MoF delivers stronger personalization performance while maintaining a more scalable and parameter-efficient design than prior approaches.
  Additional analysis indicates strong generalization to unseen users.
\end{abstract}

\section{Introduction}
Large Language Models (LLMs) have demonstrated strong performance across a wide range of tasks, from general natural language processing to complex reasoning~\cite{openai2022chatgpt, OpenAI2024gpt, google@gemini25}. 
As LLM-based personal assistants become increasingly widespread~\cite{mok-etal-2025-exploring, li-etal-2025-hello, anthropic2026claudecode}, there has been growing interest in \textit{LLM personalization}, i.e., tailoring model responses to individual user preferences~\cite{zhang2025personalization, salemi-etal-2024-lamp}. 
Such personalization is motivated by the observation that users may prefer different responses to the same query depending on their preferences~\cite{Kirk2024benebitpersonalizing, kirk2024prism, chen2024large}. 
In this work, we view user preference as a relative tendency to favor one response over another for a given query and aim to steer LLM outputs toward responses that better align with such preferences.

Despite their strong capabilities, most LLMs are trained using a \textit{one-size-fits-all} paradigm through next-token prediction~\cite{NEURIPS2020GPT3, flant5} and are subsequently aligned with general human preferences during post-training~\cite{ouyang2022training}.
As a result, LLMs tend to generate outputs that reflect dominant or average preferences, which creates a fundamental challenge in aligning LLM outputs with diverse human preferences~\cite{jang2023personalized, ValueKaleidoscope, feng-etal-2024-modular, zhou-etal-2024-beyond}.

To address this challenge, prior work has explored parameter-efficient fine-tuning (PEFT) approaches, such as training user-specific LoRA~\cite{hu2022lora} modules~\cite{tan-etal-2024-democratizing, qi2024FDLoRA} or incorporating user embeddings into model inputs~\cite{liu-etal-2025-llms, qiu-etal-2025-latent}.
While effective, these methods require direct access to model parameters and are therefore not applicable to \textit{black-box LLMs}, which are increasingly deployed through APIs~\cite{sun24p@BBoxAdapter}. 
To enable personalization for black-box settings, recent work has explored retrieval- and prompting-based approaches~\cite{salemi-etal-2024-lamp, richardson2023integrating, shi2025retrieval}, which avoid model adaptation but remain vulnerable to irrelevant information.
To provide more robust personalization, HYDRA~\cite{zhuang2024hydra} introduces a \textbf{reranker \& adapter} framework that personalizes black-box LLM outputs by selecting useful historical items and ranking candidate responses according to user preferences.
However, HYDRA relies on user-specific scoring heads, causing the number of personalized parameters to grow linearly with the number of users and requiring additional adaptation for unseen users. 
These limitations become increasingly problematic in practical deployments with large user populations.

To overcome these limitations, we propose \textbf{Mixture-of-Facets (MoF)}, a preference-aware mixture architecture inspired by the Mixture-of-Experts (MoE) framework~\cite{jacobs1991adaptive, shazeer2017moe, fedus2022switch}. 
Rather than modeling user preferences through user-specific parameters, MoF assumes that user preferences can be represented as compositions of shared latent preference facets. 
Based on this idea, MoF shifts from user-specific scoring heads to shared facet heads, with personalization performed through history-conditioned routing.
To encourage the router to focus on preference-related signals, MoF leverages a sparse autoencoder (SAE)~\cite{huben2024sparse, qiu-etal-2025-latent} over user histories.
This design allows scalable personalization with a fixed number of shared parameters and generalizes to users unseen during training without additional parameter updates.
To further promote facet specialization, we introduce a cluster-based sampling (CBS) strategy that exposes facet heads to diverse preference signals during training. 

Experimental results across multiple personalization tasks demonstrate that MoF achieves the best overall performance in terms of average rank, while improving scalability and parameter efficiency.
Additional analysis shows that MoF maintains competitive performance even without additional adaptation, indicating strong generalization to unseen users.
Our key contributions are summarized as follows:
\begin{itemize}
\item We propose MoF, a scalable black-box LLM personalization framework that models user preferences as compositions of shared latent preference facets.
\item We introduce a history-conditioned routing mechanism that enables personalization for unseen users without requiring user-specific adaptation.
\item We analyze the generalization and scalability of MoF, showing competitive performance on unseen users and constant personalization-parameter storage overhead as the number of users increases.
\end{itemize}

\begin{figure*}[ht]
\centering
\includegraphics[width=\textwidth]{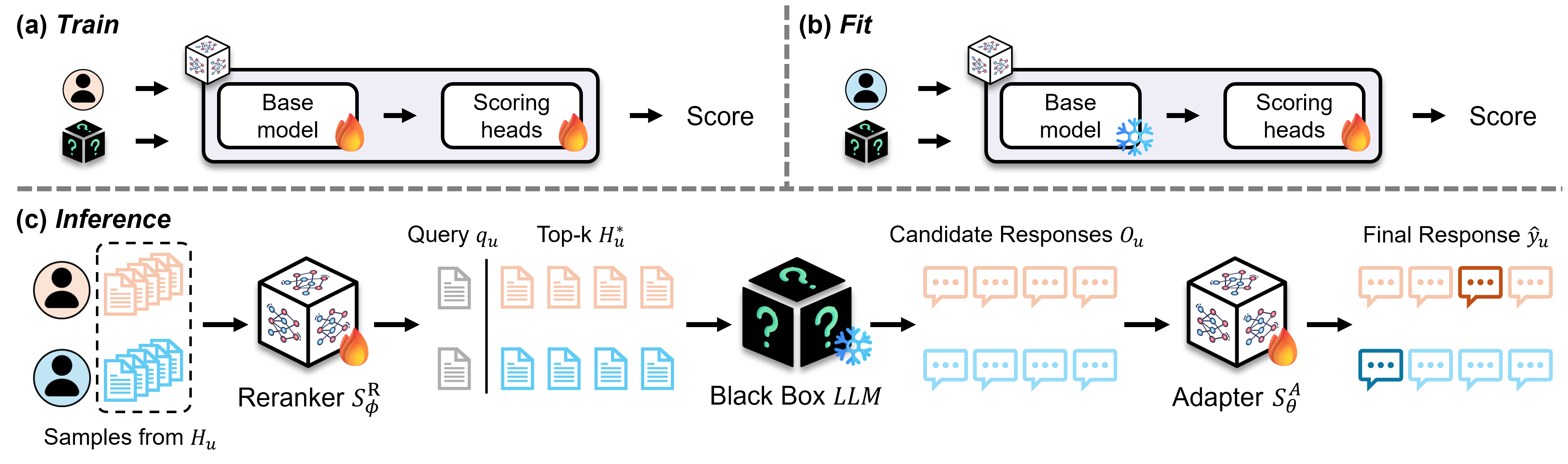}
\caption{
  Overview of the training and inference pipeline.
  \textbf{(a) \textit{Train stage}:} the shared base model and scoring heads are jointly trained using data collected from multiple users.
  \textbf{(b) \textit{Fit stage}:} the base model is frozen while only the scoring heads are updated for unseen users.
  \textbf{(c) \textit{Inference stage}:} the reranker selects useful history items, the black-box LLM generates candidate responses, and the adapter scores the candidates to produce the final response.
  }
\label{fig:ProcessOverview}
\end{figure*}
  
\section{Related Work}
\paragraph{White-Box LLM Personalization.}
LLM personalization is defined as the process of adapting a model's responses to individual users based on their profiles, characteristics, preferences, and interaction history~\cite{zhang2025personalization}.
To achieve this, PEFT-based personalization methods commonly utilize techniques such as LoRA~\cite{hu2022lora}.
OPPU~\cite{tan-etal-2024-democratizing} assigns each user a personalized LoRA module to capture user-specific preferences.
PROPER~\cite{zhang-etal-2025-proper} further improves personalization by combining parameter composition with MoE-style routing.
However, maintaining separate adapters for individual users can introduce substantial parameter and storage overhead.
To improve parameter sharing, PER-PCS~\cite{tan-etal-2024-personalized} decomposes personalized PEFT modules into reusable parameter pieces and constructs user-specific adapters by composing components from a shared parameter pool.

Although PROPER and PER-PCS employ compositional parameterization, their components differ from the preference facets considered in MoF.
PROPER models group-level user adaptations, while PER-PCS composes reusable PEFT parameter pieces to construct personalized adapters. 
In contrast, MoF represents individual preferences as weighted compositions of jointly learned shared latent preference facets, with routing weights derived from user history. 
More importantly, both PROPER and PER-PCS require access to backbone parameters and are thus limited to white-box settings, whereas MoF operates without such access and is applicable to black-box LLMs.

\paragraph{Black-Box LLM Personalization.}
To address this challenge, prior work has explored prompting-based personalization methods.
LaMP~\cite{salemi-etal-2024-lamp} introduces retrieval augmentation for personalization by retrieving personal items from user profiles.
CFRAG~\cite{shi2025retrieval} extends this direction by incorporating collaborative filtering and retrieving histories from similar users.
PAG~\cite{richardson2023integrating} combines retrieval with task-aware user summaries generated by LLMs to improve personalization.
Fermi~\cite{kim-yang-2025-shot} learns personalized prompts for each user by iteratively refining prompts using user profiles.
Although these methods improve personalization without modifying model parameters, their effectiveness can be affected by irrelevant or noisy contextual signals.

To further advance black-box LLM personalization, HYDRA~\cite{zhuang2024hydra} reranks user histories and selects preference-aligned responses via a personalized adapter.
However, HYDRA relies on user-specific scoring heads (Fig.~\ref{fig:MoFStructure} (a)), causing personalized parameters to grow linearly with the number of users and requiring an additional adaptation stage for unseen users (Fig.~\ref{fig:ProcessOverview} (b)).
In contrast, MoF enables scalable personalization through shared facet heads and history-conditioned routing, supporting personalization for unseen users without additional adaptation while requiring no dedicated user-specific parameters.

\section{Methodology}

\subsection{Preliminary}
Given a user set $\mathcal{U}=\{u_1,\dots,u_N\}$, the goal of LLM personalization is to generate a response $\hat{y}_u$ aligned with user preferences for a query $q_u$.
For each user $u$, the behavior history is defined as $H_u=\{h_u^1,\dots,h_u^T\}$, where each history item $h_u^t=(q_u^t,y_u^t)$ consists of a past query $q_u^t$ and the corresponding user-preferred response $y_u^t$.
Since different users may prefer different responses to the same query, personalized response generation requires leveraging user-specific preference information extracted from historical interactions.

Following prior work~\cite{zhuang2024hydra}, we adopt a reranker \& adapter personalization framework for black-box LLMs.
Given a target query $q_u$, relevant history items $H'_u=\{h_{u,1}^{\prime},\dots,h_{u,M}^{\prime}\}$ are retrieved from $H_u$ and then reranked by a \textbf{reranker} $s_\phi^{\mathrm{R}}$ to construct a personalized context set $H_u^*$.
Conditioned on $H_u^*$, the black-box LLM generates a set of candidate responses $O_u=\{o_u^1,\dots,o_u^B\}$.
An \textbf{adapter} $s_\theta^{\mathrm{A}}$ scores the $O_u$ according to their alignment with user preferences, and the highest-scoring response is selected as the final output $\hat{y}_u$.
An overview of the framework is illustrated in Figure~\ref{fig:ProcessOverview} (c).
Detailed formulations of the reranker and adapter objectives are provided in Appendix~\ref{app:preliminary_details}.

\subsection{Mixture-of-Facets (MoF)}
Building on the reranker \& adapter framework introduced above, we apply \textbf{Mixture-of-Facets (MoF)} to both the reranker and adapter (Fig.~\ref{fig:MoFStructure}).
MoF consists of two key components: (1) the Facet Heads, which estimate facet-specific scores (§\ref{sec:FacetHeads}), and (2) the Preference-Aware Router, which computes routing weights based on user preferences (§\ref{sec:PreferenceAwareRouter}).
To encourage facet specialization during training, we further introduce a cluster-based sampling (CBS) strategy (§\ref{sec:ClusterBasedSampling}).

\subsubsection{Facet Heads}
\label{sec:FacetHeads}
\begin{figure}[t]
\centering
\includegraphics[width=\columnwidth]{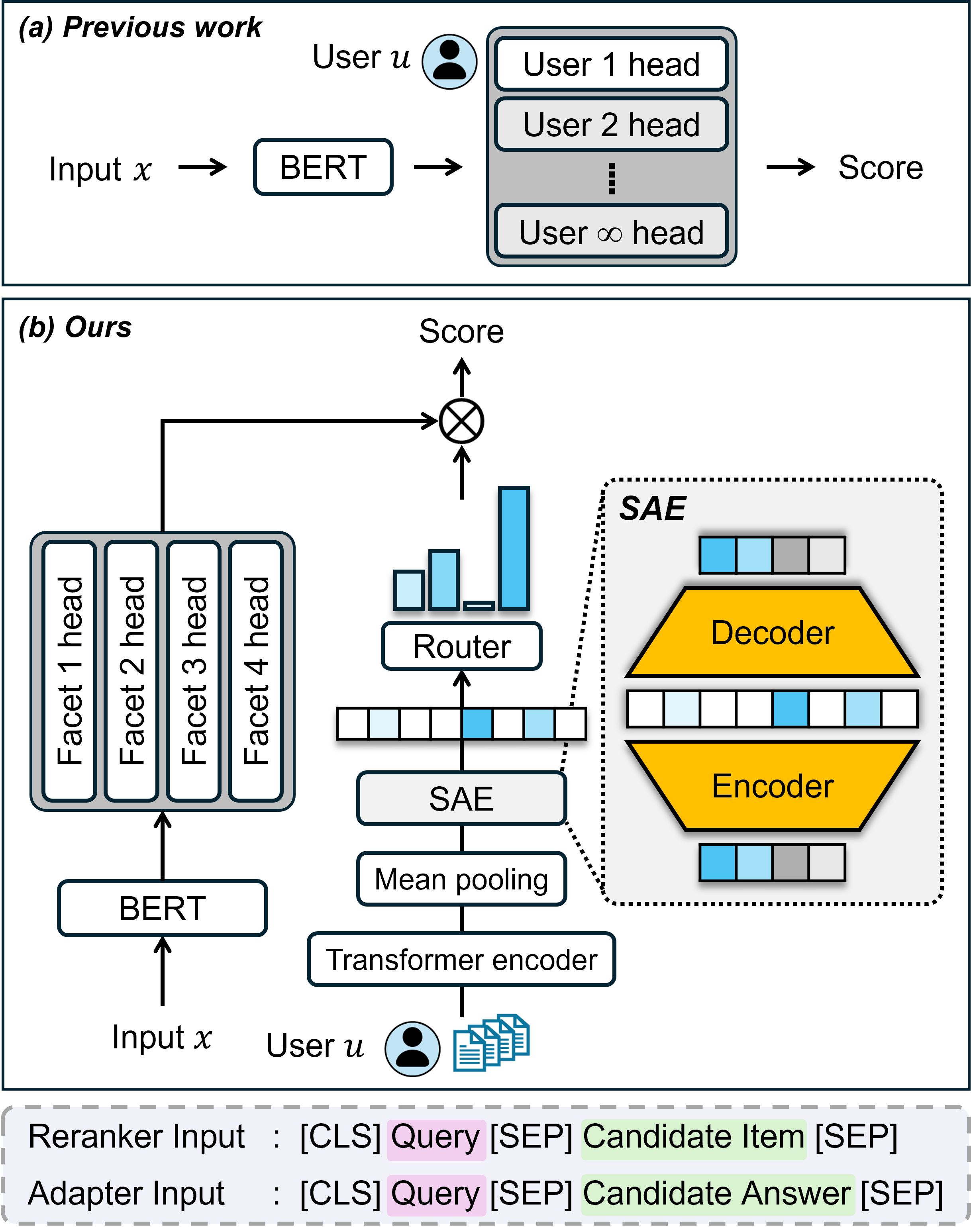}
\caption{
Comparison between (a) user-specific head personalization and (b) the proposed MoF framework.
Unlike previous work, MoF combines shared facet heads through preference-aware routing.
}
\label{fig:MoFStructure}
\end{figure}

Each facet head is implemented as a two-layer perceptron.
We intentionally use lightweight two-layer MLPs to isolate the effect of shared-facet composition while preserving parameter and computational efficiency.
Given an input pair $x^R=(q_u,h_{u,m}^{\prime})$ for the reranker or $x^A=(q_u,o_u^b)$ for the adapter, each facet head produces a preference logit.
As illustrated in Figure~\ref{fig:MoFStructure} (b), the input $x$ is first encoded by a BERT~\cite{devlin2019bert}-based encoder $\delta$, which is jointly trained with the facet heads.
We denote the final hidden representation of the [CLS] token as $\mathbf{z}_x = \delta(x)_{\texttt{[CLS]}}$.
Each facet head $\psi_f$ takes $\mathbf{z}_x$ as input and computes the corresponding preference logit:
\begin{equation}
\psi_f(\mathbf{z}_x)
=
\mathbf{W}_f^{(2)}
\,\tanh\!\left(
\mathbf{W}_f^{(1)}\mathbf{z}_x
+
\mathbf{b}_f^{(1)}
\right)
+
\mathbf{b}_f^{(2)}.
\end{equation}
Here, $\mathbf{W}_f^{(1)} \in \mathbb{R}^{d \times d}$ and $\mathbf{W}_f^{(2)} \in \mathbb{R}^{o \times d}$ denote the learnable weight matrices, where $d$ is the hidden representation dimension and $o$ denotes the dimension of the output logits.

The model contains a total of $F$ facet heads, each encouraged to specialize in different latent preference patterns during training.
The outputs of all facet heads are aggregated into the following facet logit matrix:
\begin{equation}
\Psi(x)
=
[\psi_1(\mathbf{z}_x), \dots, \psi_F(\mathbf{z}_x)]
\in
\mathbb{R}^{F \times o}.
\end{equation}
Each row of $\Psi(x)$ represents the output logits generated by a corresponding facet head.

\subsubsection{Sparse User Representation}
Independently of computing $\Psi(x)$, the router constructs a user-specific preference representation from $H_u$ in order to estimate routing weights for aggregating $\Psi(x)$.
We intentionally exclude query representations from the router input so that routing decisions focus on persistent user preference characteristics rather than query-specific semantics.
Each history item is encoded using the shared encoder $\delta$, which is kept frozen during router training.
For each history item $h_u^t$, we obtain the corresponding history embedding as $\mathbf{e}_u^t=\delta(h_u^t)_{\texttt{[CLS]}}$.
The resulting historical embedding set for user $u$ is defined as $\mathbf{E}_u=\{\mathbf{e}_u^1,\dots,\mathbf{e}_u^T\}$.
To capture patterns within the user history, we apply a Transformer encoder over $\mathbf{E}_u$ and compute the user representation via mean pooling:
\begin{equation}
\mathbf{e}^{\prime}_u
=
\mathrm{Mean}
\left(
\mathrm{Encoder}(\mathbf{E}_u)
\right).
\end{equation}

The dense user representation $\mathbf{e}^{\prime}_u$ contains general semantic and syntactic information that is not directly relevant to user preferences.
To facilitate the router's focus on preference-related signals, we further transform the dense representation into a sparse latent representation using a Sparse Autoencoder (SAE)~\cite{huben2024sparse}.
The SAE projects the dense user representation into a higher-dimensional sparse latent space, encouraging preference-related activation patterns to be selectively represented across different latent dimensions.
The sparse latent representation is computed as
\begin{equation}
\mathbf{e}^{enc}_u
=
\mathrm{ReLU}
\left(
\mathbf{W}_{\mathrm{enc}}\mathbf{e}^{\prime}_u
+
\mathbf{b}_{\mathrm{enc}}
\right),
\end{equation}
where $\mathbf{e}^{enc}_u \in \mathbb{R}^{\tau d}$ and $\mathbf{W}_{\mathrm{enc}} \in \mathbb{R}^{\tau d \times d}$.
Here, $\tau$ denotes the latent expansion factor that controls the dimensionality of the sparse representation.
This latent space provides additional capacity for sparse activations, facilitating the selective representation of preference-related signals.
The resulting sparse representation $\mathbf{e}^{enc}_u$ is used as the router input representation.

The SAE decoder reconstructs the original dense representation from the sparse latent representation:
\begin{equation}
\mathbf{e}^{dec}_u
=
\mathbf{W}_{\mathrm{dec}}\mathbf{e}^{enc}_u
+
\mathbf{b}_{\mathrm{dec}},
\end{equation}
where $\mathbf{e}^{dec}_u \in \mathbb{R}^{d}$ and $\mathbf{W}_{\mathrm{dec}} \in \mathbb{R}^{d \times \tau d}$.
The SAE is optimized using the following objective:
\begin{equation}
\mathcal{L}_{\mathrm{SAE}}
=
\left\|
\mathbf{e}^{dec}_u
-
\mathbf{e}^{\prime}_u
\right\|_2^2
+
\lambda_1
\left\|
\mathbf{e}^{enc}_u
\right\|_1.
\end{equation}
The reconstruction objective preserves preference-relevant information encoded in the user-history representation, while the \(L_1\) regularization promotes sparsity in the latent representation.

\subsubsection{User Preference-Aware Router}
\label{sec:PreferenceAwareRouter}
Using the sparse user representation $\mathbf{e}^{enc}_u$, the router determines which facet heads are more relevant for a particular user.
The router is implemented as a linear projection layer that produces routing logits over the $F$ facet heads:
\begin{equation}
\mathbf{g}_u
=
\mathbf{W}_r \mathbf{e}^{enc}_u
+
\mathbf{b}_r,
\end{equation}
where $\mathbf{W}_r \in \mathbb{R}^{F \times \tau d}$ denotes the routing projection matrix and $\mathbf{g}_u \in \mathbb{R}^{F}$ represents the routing logits over all facet heads.
To encourage sparse activation of facet heads, we apply top-$k$ routing over $\mathbf{g}_u$:
\begin{equation}
(\tilde{\mathbf{g}}_u,\mathcal{K}_u)
=
\text{top-}k
(
\mathbf{g}_u,K
),
\end{equation}
where $\tilde{\mathbf{g}}_u \in \mathbb{R}^{K}$ denotes the selected top-$k$ routing logits and $\mathcal{K}_u$ represents the corresponding facet indices.
The routing weights are then computed using a softmax over the selected routing logits:
\begin{equation}
\alpha_{u,k}
=
\frac{
\exp(\tilde{g}_{u,k})
}{
\sum_{j=1}^{K}
\exp(\tilde{g}_{u,j})
}.
\end{equation}
The final MoF prediction is obtained as a weighted sum of the selected facet outputs using the routing weights:
\begin{equation}
\mathrm{MoF}(x,H_u)
=
\sum_{k=1}^{K}
\alpha_{u,k}
\,
\psi_{\mathcal{K}_{u,k}}(x).
\end{equation}
This sparse routing mechanism enables the model to represent user-specific preference patterns as compositions of shared facet heads, allowing personalization without dedicated user-specific heads.

\subsubsection{Objective}
The model is trained using a binary classification objective over the MoF prediction and the corresponding supervision label $r_i^u$.
The labeling strategy is described in Appendix~\ref{app:preliminary_details}.
For a user $u$ and training instance $(x_i^u,r_i^u)$, the prediction probability is computed as
\begin{equation}
p_i^u
=
\sigma(\mathrm{MoF}(x_i^u,H_u)).
\end{equation}
The MoF objective is defined as
\begin{equation}
\mathcal{L}_{\mathrm{MoF}}
=
-r_i^u \log p_i^u
-
(1-r_i^u)\log(1-p_i^u).
\end{equation}
The final training objective jointly optimizes the MoF objective and the sparse autoencoder objective:
\begin{equation}
\mathcal{L}
=
\mathcal{L}_{\mathrm{MoF}}
+
\lambda
\mathcal{L}_{\mathrm{SAE}}.
\end{equation}
Here, $\lambda$ controls the contribution of the SAE regularization objective during training.

\subsection{Cluster-Based Sampling (CBS)}
\label{sec:ClusterBasedSampling}
In HYDRA, training instances are constructed by retrieving history items based on relevance to the target query.
However, retrieval relevance often emphasizes lexical or semantic similarity, which may fail to reflect the diverse aspects of user preferences.
As a result, the sampled training instances may be concentrated on a limited subset of preference patterns.
To encourage facet specialization in MoF, we introduce a cluster-based sampling (CBS) strategy over $H_u$.
Specifically, we first partition history items into multiple clusters using k-means clustering over the history embedding set $\mathbf{E}_u$.
CBS is applied when constructing training instances for both the reranker and adapter.
Rather than sampling history items from a single global retrieval pool, CBS samples them separately from each cluster.
This strategy increases the diversity of preference signals observed during training and encourages the facet heads to capture diverse preference patterns rather than focusing on semantically similar history items.

\section{Experiment}
\begin{table*}[t]
\caption{
Main results on the LaMP benchmark.
Results are averaged over three runs with different sampled user sets.
$k$ denotes the number of retrieved user-history items.
For HYDRA and MoF, we use $k=4$.
$\uparrow$ and $\downarrow$ indicate that higher and lower values are better, respectively.
\textbf{Bold} and \underline{underline} indicate the best and second-best scores, respectively.
Avg. Rank denotes the average rank of each method across all metrics.
}
\centering
\small
\setlength{\tabcolsep}{4pt}
\renewcommand{\arraystretch}{1.1}
\resizebox{\textwidth}{!}{
\begin{tabular}{l cc cc ccc ccc c}
\toprule
\textbf{Dataset ($\rightarrow$)} 
  & \multicolumn{2}{c}{\textbf{LaMP-2}} 
  & \multicolumn{2}{c}{\textbf{LaMP-3}} 
  & \multicolumn{3}{c}{\textbf{LaMP-4}} 
  & \multicolumn{3}{c}{\textbf{LaMP-5}}
  & \multirow{2}{*}{\shortstack{\textbf{Avg.}\\\textbf{Rank}$\downarrow$}} \\
\cmidrule(lr){2-3} \cmidrule(lr){4-5} \cmidrule(lr){6-8} \cmidrule(lr){9-11}
\textbf{Method ($\downarrow$)} 
  & Acc~$\uparrow$ & F1~$\uparrow$ 
  & MAE~$\downarrow$ & RMSE~$\downarrow$ 
  & R-1~$\uparrow$ & R-L~$\uparrow$ & BLEU~$\uparrow$ 
  & R-1~$\uparrow$ & R-L~$\uparrow$ & BLEU~$\uparrow$ 
  & \\
\midrule

Zero-Shot
& 0.420 & 0.257 
& 0.500 & 0.777 
& 0.147 & 0.129 & 0.891
& 0.446 & 0.375 & 5.822
& 13.5 \\

\midrule

ICL ($k$=1)
& 0.393 & 0.252 
& 0.727 & 1.163 
& 0.158 & 0.140 & 0.977
& 0.444 & 0.373 & 6.098
& 13.6 \\

ICL ($k$=2)
& 0.467 & 0.387 
& 0.520 & 0.879 
& 0.167 & 0.143 & 0.837
& 0.448 & 0.381 & 6.032
& 11.3 \\

ICL ($k$=4)
& 0.513 & 0.403 
& 0.433 & 0.762 
& 0.161 & 0.140 & 1.240
& 0.447 & 0.384 & 5.511
& 10.0 \\

\midrule

RAG ($k$=1)
& 0.433 & 0.276 
& 0.613 & 0.978 
& 0.165 & 0.145 & 0.889
& 0.452 & 0.385 & 5.961
& 12.0 \\

RAG ($k$=2)
& 0.493 & 0.346 
& 0.500 & 0.812 
& 0.167 & 0.150 & 1.273
& 0.459 & 0.387 & 5.388
& 9.5 \\

RAG ($k$=4)
& 0.547 & 0.409 
& 0.440 & 0.765 
& 0.176 & 0.148 & 1.456
& 0.457 & 0.395 & 5.888
& 6.9 \\

\midrule

PAG ($k$=0)
& 0.507 & 0.366 
& 0.333 & \underline{0.663} 
& 0.162 & 0.140 & 0.873
& 0.436 & 0.361 & 5.659
& 10.5 \\

PAG ($k$=1)
& 0.473 & 0.337 
& 0.400 & 0.739 
& 0.181 & 0.157 & 1.040
& 0.454 & 0.371 & 6.063
& 8.0 \\

\midrule

CFRAG ($k$=4)
& 0.593 &	\underline{0.447} 
&	0.440 &	0.780 
&	0.176 &	0.151 &	0.966 
&	0.456 &	0.392 &	5.748
& 7.4 \\

\midrule

HYDRA-Reranker
& 0.567 & 0.383 
& 0.480 & 0.830 
& 0.168 & 0.152 & 1.517
& 0.465 & 0.390 & 6.281
& 6.8 \\

HYDRA-Adapter
& 0.447 & 0.271 
& 0.380 & 0.735 
& 0.159 & 0.140 & 0.852
& 0.467 & 0.394 & 6.059
& 9.3 \\

HYDRA
& 0.593 & 0.417 
& 0.393 & 0.761 
& \underline{0.183} & \underline{0.163} & \textbf{1.849}
& 0.459 & 0.396 & 6.652
& \underline{3.5} \\

\midrule

\rowcolor{gray!15}
MoF-Reranker (ours)
& \underline{0.600} & \textbf{0.451} 
& 0.407 & 0.724 
& 0.177 & 0.154 & 1.502
& 0.456 & 0.390 & \textbf{7.421}
& 4.2 \\

\rowcolor{gray!15}
MoF-Adapter (ours)
& 0.467 & 0.308 
& \underline{0.327} & 0.682 
& 0.161 & 0.142 & 1.275
& \underline{0.477} & \underline{0.400} & 6.515
& 6.5 \\

\rowcolor{gray!30}
MoF (ours)
& \textbf{0.627} & 0.438 
& \textbf{0.300} & \textbf{0.605} 
& \textbf{0.199} & \textbf{0.176} & \underline{1.662}
& \textbf{0.481} & \textbf{0.413} & \underline{6.967}
& \textbf{1.4} \\

\bottomrule
\end{tabular}
}
\label{tab:MainResults}
\end{table*}
\subsection{Experimental Setup}
\paragraph{Datasets.}
We evaluate MoF on two personalization classification tasks and two personalization generation tasks from the LaMP benchmark~\cite{salemi-etal-2024-lamp}. 
Specifically, we use \textbf{Movie Tagging Classification (LaMP-2)} and \textbf{Product Rating Classification (LaMP-3)} for classification, and \textbf{News Headline Generation (LaMP-4)} and \textbf{Scholarly Title Generation (LaMP-5)} for generation.
Further details on the datasets are provided in Appendix~\ref{app:taskdetails}.
\paragraph{Baselines.}
We compare MoF against both prompt-based and reranker \& adapter-based personalization methods.
\textbf{Zero-Shot} uses only the input query without any personalization cues.
\textbf{ICL} augments the input with randomly selected $k$ items from the user history.
\textbf{RAG}~\cite{salemi-etal-2024-lamp} retrieves relevant history items using BM25~\cite{robertson2009probabilistic} and augments them into the input.
\textbf{PAG}~\cite{richardson2023integrating} further enhances personalization by incorporating LLM-generated user profile summaries together with retrieved history items.
\textbf{CFRAG}~\cite{shi2025retrieval} incorporates collaborative information by retrieving similar users and relevant items from their histories, followed by personalized reranking for LLM generation.
\textbf{HYDRA}~\cite{zhuang2024hydra} is a reranker \& adapter-based personalization framework.
We additionally report \textbf{HYDRA-Reranker}, which uses only the reranker for personalization, and \textbf{HYDRA-Adapter}, which applies only the adapter over candidate responses generated from the zero-shot setting.
\paragraph{Evaluation Metrics.}
Following LaMP~\cite{salemi-etal-2024-lamp}, we use Accuracy (Acc) and F1 score (F1) for LaMP-2, MAE and RMSE for LaMP-3, and ROUGE-1 (R-1), ROUGE-L (R-L), and BLEU for LaMP-4 and LaMP-5.
Higher scores indicate better performance for all metrics except MAE and RMSE.
\paragraph{Implementation Details.}
Following prior work~\cite{zhuang2024hydra}, we conduct training and evaluation using subsets of the original datasets, where we sample 100 users for training and 50 users for testing. 
For robustness, results are averaged over 3 runs with different randomly sampled user sets. 
We use \texttt{bge-base-en-v1.5}~\cite{bgebaseen} (110M parameters) as the base model for both the reranker and adapter modules.
As the black-box LLM, we employ \texttt{gpt-3.5-turbo-1106}~\cite{openai2022chatgpt}. 
Additional implementation details are provided in Appendix~\ref{app:ImplementationDetails}.

\subsection{Main Results}
Table~\ref{tab:MainResults} presents the main experimental results on the LaMP benchmark.
We observe that prompt-based personalization methods do not consistently improve over the zero-shot setting and in some cases even degrade performance.
One possible explanation is that prompt-based methods are sensitive to history selection and context coverage because randomly selected or limited histories may fail to represent the user's preferences while introducing distracting context. 
The extent of this effect also varies across tasks and retrieval sizes.
In contrast, reranker \& adapter-based approaches generally provide more consistent gains across tasks, suggesting that aligning the output distribution of a black-box LLM with user preferences can be more effective than relying solely on prompt-level personalization.

Compared with HYDRA, which relies on user-specific scoring heads, MoF achieves better overall performance despite eliminating user-specific parameters.
For example, MoF improves accuracy from 0.593 to 0.627 on LaMP-2 and reduces MAE from 0.393 to 0.300 on LaMP-3.

MoF achieves the strongest average performance across the tasks with an average rank of 1.4, demonstrating stronger overall performance than existing prompt-based and reranker \& adapter-based baselines.
Specifically, MoF achieves the best performance on most evaluation metrics across LaMP-2, LaMP-3, LaMP-4, and LaMP-5, with competitive results on the remaining metrics.
These results demonstrate the effectiveness of preference-aware mixture modeling for black-box LLM personalization.

We further observe that both MoF-Reranker and MoF-Adapter individually achieve competitive performance. 
This suggests that preference-aware retrieval and personalized candidate selection provide complementary benefits.
Additional experimental results on other black-box LLMs are provided in Appendix~\ref{app:GeneralizationToOtherModels}, suggesting that the proposed framework generalizes across different proprietary black-box LLMs.

\subsection{Generalization to Unseen Users}
\label{sec:GeneralizationtoUnseenUsers}
Unlike HYDRA, which requires an additional fit stage for unseen users because of its user-specific heads, MoF can perform personalization without additional training for new users. 
Table~\ref{tab:PersonalizationForUnseenUsers} presents the generalization performance on unseen users with and without the fit stage.
MoF maintains competitive performance even without the fit stage, with only minor degradation across datasets.
Notably, MoF without the fit stage outperforms HYDRA with the fit stage on all metrics reported in Table~\ref{tab:PersonalizationForUnseenUsers}, demonstrating strong generalization to unseen users.
While MoF-Adapter does not outperform its fitted counterpart in every case, it remains competitive without additional training.
These findings suggest that MoF can effectively generalize to unseen users without requiring user-specific adaptation.
\begin{table}[t]
\caption{
Evaluation of unseen user generalization: performance with and without the fit stage.
\textbf{--} indicates that the method cannot be applied without a fit stage.
}
\centering
\small
\setlength{\tabcolsep}{4pt}
\renewcommand{\arraystretch}{1.0}
\newcommand{\cmark}{\textcolor{green!70!black}{\ding{51}}}
\newcommand{\xmark}{\textcolor{red}{\ding{55}}}
\resizebox{\columnwidth}{!}{
\begin{tabular}{l c c c cc}
\toprule
\textbf{Dataset ($\rightarrow$)} &
  & \multicolumn{1}{c}{\textbf{\footnotesize LaMP-2}} 
  & \multicolumn{1}{c}{\textbf{\footnotesize LaMP-3}} 
  & \multicolumn{2}{c}{\textbf{\footnotesize LaMP-4}}  \\
\cmidrule(lr){3-3} \cmidrule(lr){4-4} \cmidrule(lr){5-6}
\textbf{Method ($\downarrow$)} 
  & \textbf{Fit}
  & Acc~$\uparrow$
  & MAE~$\downarrow$
  & R-1~$\uparrow$ & R-L~$\uparrow$ \\

\midrule

HYDRA & \xmark
& \textbf{--}
& \textbf{--}
& \textbf{--} & \textbf{--} \\

HYDRA & \cmark
& 0.593
& 0.393
& 0.183 & 0.163 \\

\rowcolor{gray!30}
MoF & \xmark
& \underline{0.620}
& \underline{0.333}
& \textbf{0.200} & \textbf{0.179} \\

\rowcolor{gray!30}
MoF & \cmark
& \textbf{0.627}
& \textbf{0.300}
& \underline{0.199} & \underline{0.176} \\

\midrule

HYDRA-Reranker & \xmark
& \textbf{--}
& \textbf{--}
& \textbf{--} & \textbf{--} \\

HYDRA-Reranker & \cmark
& \underline{0.567}
& 0.480
& 0.168 & 0.152 \\

\rowcolor{gray!15}
MoF-Reranker & \xmark
& 0.547
& \underline{0.440}
& \textbf{0.178} & \textbf{0.158} \\

\rowcolor{gray!15}
MoF-Reranker & \cmark
& \textbf{0.600}
& \textbf{0.407}
& \underline{0.177} & \underline{0.154} \\

\midrule

HYDRA-Adapter & \xmark
& \textbf{--}
& \textbf{--}
& \textbf{--} & \textbf{--} \\

HYDRA-Adapter & \cmark
& \underline{0.447}
& 0.380
& \underline{0.159} & \underline{0.140} \\

\rowcolor{gray!15}
MoF-Adapter & \xmark
& 0.440
& \underline{0.347}
& 0.153 & 0.137 \\

\rowcolor{gray!15}
MoF-Adapter & \cmark
& \textbf{0.467}
& \textbf{0.327}
& \textbf{0.161} & \textbf{0.142} \\

\bottomrule

\end{tabular}
}
\label{tab:PersonalizationForUnseenUsers}
\end{table}

\begin{figure}[t]
  \centering
\includegraphics[width=\columnwidth]{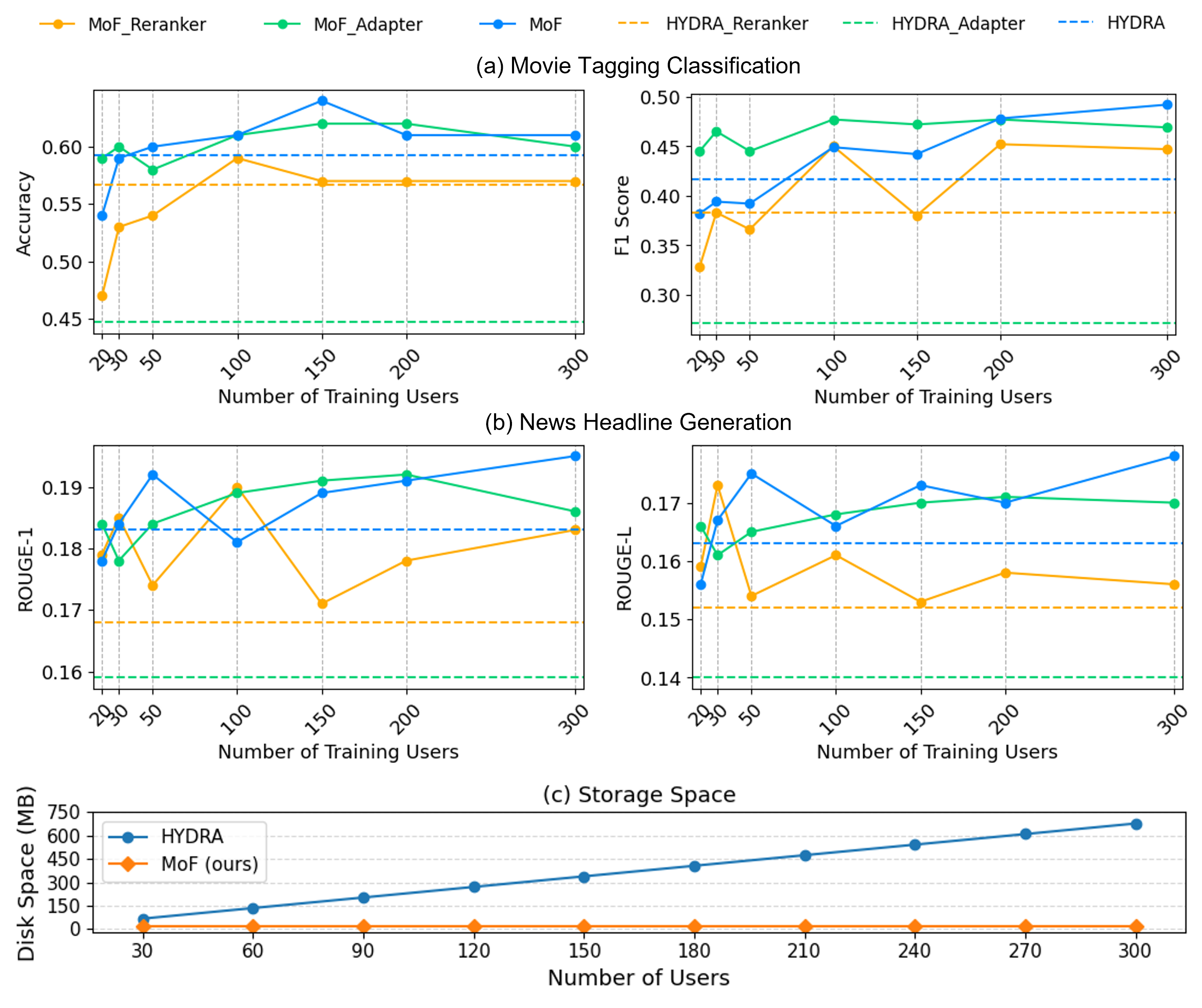}
\caption{Scalability analysis with increasing user populations.
(a), (b) show personalization performance as the number of training users increases, while (c) compares storage growth across methods.}
\label{fig:NumberofTrainingUserAnal}
\end{figure}

\begin{table}[t]
\caption{
  Large-scale evaluation on 1,000 unseen users without user-specific fitting.
  }
\centering
\small
\setlength{\tabcolsep}{5.5pt}
\renewcommand{\arraystretch}{1.0}

\begin{tabular}{l c c cc}
\toprule
\textbf{Dataset ($\rightarrow$)} 
  & \multicolumn{1}{c}{\textbf{LaMP-2}} 
  & \multicolumn{1}{c}{\textbf{LaMP-3}} 
  & \multicolumn{2}{c}{\textbf{LaMP-4}}  \\
\cmidrule(lr){2-2} \cmidrule(lr){3-3} \cmidrule(lr){4-5}
\textbf{Method ($\downarrow$)} 
  & Acc~$\uparrow$
  & MAE~$\downarrow$
  & R-1~$\uparrow$ & R-L~$\uparrow$ \\
\midrule

Zero-Shot
& 0.391
& 0.515
& 0.143 & 0.124 \\

\midrule

RAG ($k=1$)
& 0.409
& 0.676
& 0.155 & 0.135 \\

RAG ($k=2$)
& 0.480 
& 0.521 
& 0.160 & 0.141 \\

RAG ($k=4$)
& \underline{0.508}
& \underline{0.449}
& \underline{0.164} & \underline{0.146} \\

\midrule

\rowcolor{gray!30}
MoF (ours)
& \textbf{0.577}
& \textbf{0.354}
& \textbf{0.175} & \textbf{0.156} \\

\bottomrule
\end{tabular}

\label{tab:LargeScaleEval}
\end{table}
\subsection{Scalability Analysis}
Figure~\ref{fig:NumberofTrainingUserAnal} (a) and (b) show the effect of the number of training users.
Across both datasets, MoF achieves competitive performance with substantially fewer training users than HYDRA.
For example, MoF trained with only 30 users in (a) and 50 users in (b) already surpasses HYDRA trained with 100 users, suggesting that MoF can achieve strong personalization performance with fewer training users.
Furthermore, performance remains stable as the number of training users increases up to 300, suggesting that the learned shared facets remain effective as the training user population grows.

Figure~\ref{fig:NumberofTrainingUserAnal} (c) compares storage growth as the number of users increases.
While HYDRA requires user-specific scoring heads, causing storage cost to grow linearly with the user population (approximately 2.26 MB per user in our implementation), MoF maintains constant personalization-parameter storage cost through shared facet heads.
Appendix~\ref{app:ParameterScalability} further presents a theoretical analysis showing that MoF exhibits constant personalization complexity with respect to the number of users.

To assess scalability to a larger user population, we evaluate MoF, trained on 100 users, on 1,000 unseen users without additional user-specific fitting.
As shown in Table~\ref{tab:LargeScaleEval}, MoF consistently outperforms the zero-shot and RAG baselines on LaMP-2, LaMP-3, and LaMP-4.
These results provide further evidence of inference-time scalability and unseen-user generalization to a substantially larger user population.

\begin{table}[t]
\caption{
  Controlled comparison of HYDRA and MoF with and without CBS.
}
\centering
\small
\setlength{\tabcolsep}{4pt}
\renewcommand{\arraystretch}{1.0}
\newcommand{\cmark}{\textcolor{green!70!black}{\ding{51}}}
\newcommand{\xmark}{\textcolor{red}{\ding{55}}}
\resizebox{\columnwidth}{!}{
\begin{tabular}{l c c c cc}
\toprule
\textbf{Dataset ($\rightarrow$)} &
  & \multicolumn{1}{c}{\textbf{\footnotesize LaMP-2}} 
  & \multicolumn{1}{c}{\textbf{\footnotesize LaMP-3}} 
  & \multicolumn{2}{c}{\textbf{\footnotesize LaMP-4}}  \\
\cmidrule(lr){3-3} \cmidrule(lr){4-4} \cmidrule(lr){5-6}
\textbf{Method ($\downarrow$)} 
  & \textbf{CBS}
  & Acc~$\uparrow$
  & MAE~$\downarrow$
  & R-1~$\uparrow$ & R-L~$\uparrow$ \\

\midrule

HYDRA & \xmark
& \underline{0.593}
& 0.393
& \underline{0.183} & 0.163 \\

HYDRA & \cmark
& 0.560
& \underline{0.373}
& 0.170 & 0.149 \\

\rowcolor{gray!30}
MoF & \xmark
& 0.587
& 0.380
& \underline{0.183} & \underline{0.168} \\

\rowcolor{gray!30}
MoF & \cmark
& \textbf{0.627}
& \textbf{0.300}
& \textbf{0.199} & \textbf{0.176} \\

\bottomrule

\end{tabular}
}
\label{tab:ControlledAnalysisCBS}
\end{table}

\begin{table}[t]
  \caption{
    Effect of applying CBS to different components of MoF.
    R and A denote the reranker and adapter, respectively.
  }
  \centering
  \small
  \setlength{\tabcolsep}{5.5pt}
  \renewcommand{\arraystretch}{0.9}
  
  \newcommand{\cmark}{\textcolor{green!70!black}{\ding{51}}}
  \newcommand{\xmark}{\textcolor{red}{\ding{55}}}
  
  \begin{tabular}{c cc c c cc}
  \toprule
  
  & \multicolumn{2}{c}{\textbf{CBS}}
  & \multicolumn{1}{c}{\textbf{\footnotesize LaMP-2}}
  & \multicolumn{1}{c}{\textbf{\footnotesize LaMP-3}}
  & \multicolumn{2}{c}{\textbf{\footnotesize LaMP-4}} \\
  \cmidrule(lr){2-3}
  \cmidrule(lr){4-4}
  \cmidrule(lr){5-5}
  \cmidrule(lr){6-7}
  
  & \textbf{R}
  & \textbf{A}
  & Acc~$\uparrow$
  & MAE~$\downarrow$
  & R-1~$\uparrow$
  & R-L~$\uparrow$ \\
  \midrule
  
  & \xmark & \xmark
  & 0.587
  & 0.380
  & \underline{0.183}
  & \underline{0.168} \\
  
  MoF
  & \cmark & \xmark
  & \underline{0.613}
  & \underline{0.327}
  & 0.179
  & 0.155 \\

  \rowcolor{gray!30}
  & \cmark & \cmark
  & \textbf{0.627}
  & \textbf{0.300}
  & \textbf{0.199}
  & \textbf{0.176} \\
  
  \bottomrule
  \end{tabular}
  
  \label{tab:CBSAnalComponent}
  \end{table}

\subsection{Controlled Analysis of CBS}
\label{sec:cbs_analysis}
To disentangle the effect of CBS from that of the MoF architecture, we conduct controlled comparisons with and without CBS.
As shown in Table~\ref{tab:ControlledAnalysisCBS}, applying CBS to HYDRA degrades performance on most reported metrics, whereas combining CBS with MoF improves performance across the reported metrics.
Notably, MoF without CBS remains competitive with HYDRA, suggesting that CBS alone does not account for the gains of MoF.
Rather, CBS complements the shared-facet architecture by exposing facet heads to diverse preference signals, encouraging specialization across facets.

We further examine where CBS contributes within MoF by selectively applying it to the reranker and adapter.
As shown in Table~\ref{tab:CBSAnalComponent}, applying CBS to the reranker improves performance on LaMP-2 and LaMP-3 but slightly decreases the ROUGE metrics on LaMP-4. 
Extending CBS to both the reranker and adapter further improves LaMP-3 and recovers and improves the ROUGE metrics on LaMP-4.
These results suggest that extending CBS to both components can provide additional benefits within the MoF framework.

\begin{table}[t]
\caption{
Ablation study of MoF. 
% We analyze the contribution of each component by removing SAE, preference-aware router, multiple facet heads, and cluster-based sampling (CBS).
}
\centering
\small
\setlength{\tabcolsep}{4pt}
\renewcommand{\arraystretch}{1.0}
\resizebox{\columnwidth}{!}{
\begin{tabular}{l c c cc}
\toprule
\textbf{Dataset ($\rightarrow$)} 
  & \multicolumn{1}{c}{\textbf{\footnotesize LaMP-2}} 
  & \multicolumn{1}{c}{\textbf{\footnotesize LaMP-3}} 
  & \multicolumn{2}{c}{\textbf{\footnotesize LaMP-4}} \\
\cmidrule(lr){2-2} \cmidrule(lr){3-3} \cmidrule(lr){4-5}
\textbf{Method ($\downarrow$)} 
  & Acc~$\uparrow$
  & MAE~$\downarrow$
  & R-1~$\uparrow$ & R-L~$\uparrow$ \\

\midrule

\rowcolor{gray!30}
MoF
& \textbf{0.627}
& \textbf{0.300}
& \textbf{0.199} & \textbf{0.176} \\

~~ w/o SAE
& \underline{0.607}
& \underline{0.333}
& \underline{0.188} & 0.165 \\

~~ w/o Router
& 0.600
& 0.360
& 0.184 & 0.161 \\

~~ w/o Multi-Facet
& \underline{0.607}
& 0.353 
& 0.183 & 0.160 \\

~~ w/o CBS
& 0.587 
& 0.380 
& 0.183 & \underline{0.168} \\

\bottomrule
\end{tabular}
}
\label{tab:MainAblationStudy}
\end{table}

\begin{figure}[t]
  \centering
  \includegraphics[width=\columnwidth]{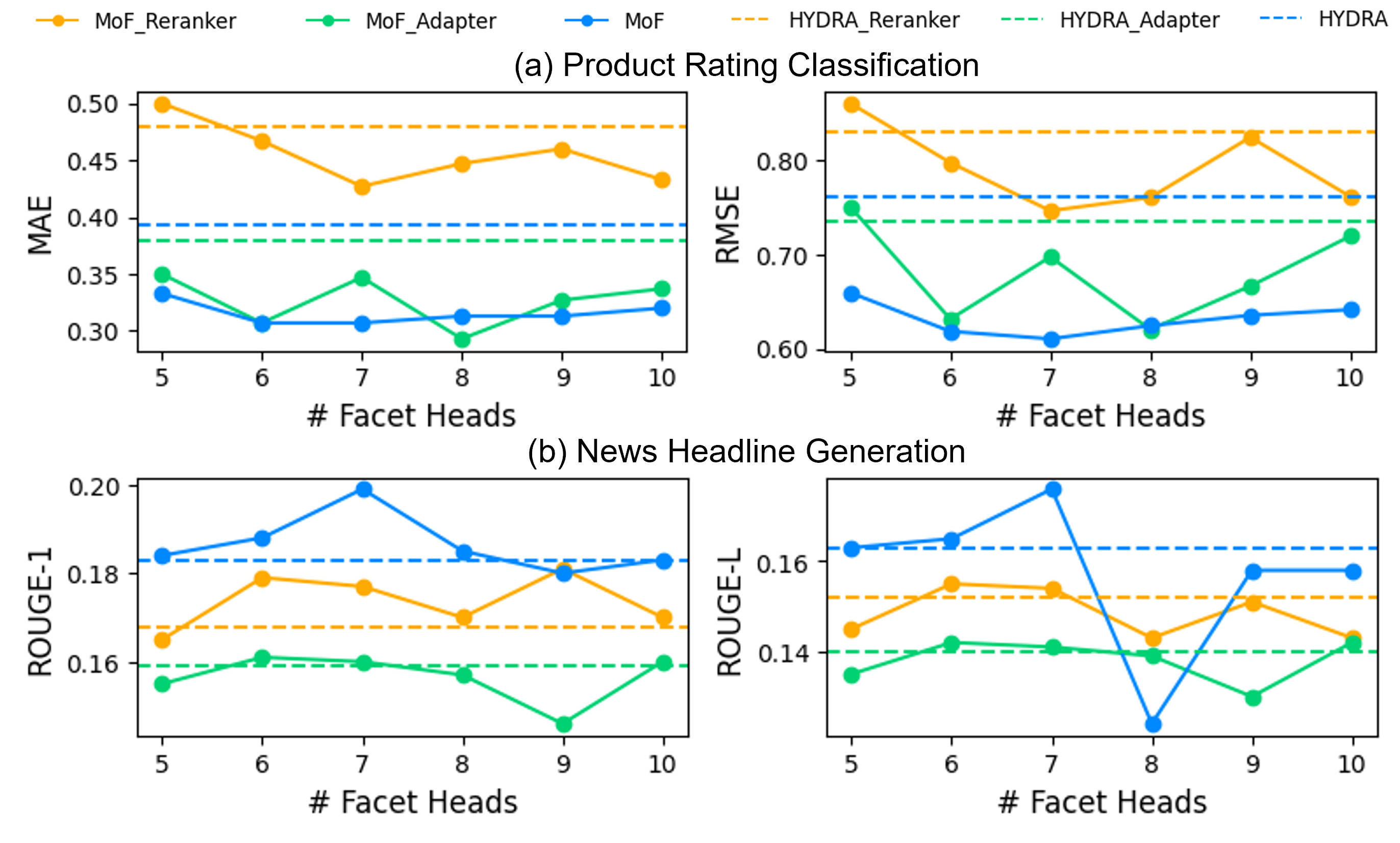}
  \caption{Impact of the number of facet heads under top-$3$ routing.
  (a) and (b) show results on LaMP-3 and LaMP-4, respectively.
  A moderate number of facet heads generally provides the strongest performance.}
  \label{fig:NumOfFacetsAnal}
\end{figure}

\subsection{Ablation Studies}
Table~\ref{tab:MainAblationStudy} presents an ablation study evaluating the contribution of each component in MoF.
Removing CBS results in the largest performance degradation in most settings, indicating the importance of exposing the model to diverse training samples during preference learning.
We also observe substantial performance drops when removing the router, demonstrating the benefit of adaptively combining shared preference facets.
Replacing multiple shared facet heads with a single head also degrades performance, suggesting that user preferences are better captured by combinations of multiple latent preference patterns.
Finally, removing SAE results in smaller but consistent performance degradation across datasets, indicating that sparse representations provide complementary benefits for personalization.

\begin{table*}[t]
\caption{
Qualitative analysis of user groups identified from routing patterns. 
Representative n-grams are obtained using BERTopic on user histories within each group.
}
\centering
\small
\setlength{\tabcolsep}{4pt}
\renewcommand{\arraystretch}{1.0}

\begin{tabular}{c c c}
\toprule

\textbf{ID} &
\textbf{Representative n-grams} &
\textbf{Interpretation}
\\

\midrule

$G1$
&
\parbox{8.8cm}{
“17th century salem hester prynne”; “years old daughter maddy”; “abuse fanatically pious mother”; “wormhole surpass limitations human space”
}
&
\parbox{5.7cm}{
Threat-, escape-, and conflict-centered narratives spanning historical, familial, supernatural, and speculative settings
}
\\

\midrule

$G2$
&
\parbox{8.8cm}{
“zion defends massive invasion machines”; “young wizards harry discovers trio”; “11th birthday learns powerful wizard”; “zombies perfectly evolved survivors”    }
&
\parbox{5.7cm}{
Genre-driven action/fantasy narratives emphasizing adventure, large-scale conflict, and survival    }
\\

\midrule

$G3$
&
\parbox{8.8cm}{
“actively opposed hitler nazis convictions”; “young prince life threatened evil”; “adventures lion witch wardrobe”; “young writer elizabeth wurtzel earns”; “suddenly single cal adrift”
}
&
\parbox{5.7cm}{
Character-centered dramatic narratives involving moral conviction, family/romance, youth vulnerability, and emotional struggle    }
\\

\bottomrule
\end{tabular}
\label{tab:GroupTopicModelingLaMP2}

\end{table*}

\subsection{Impact of Number of Facet Heads}
Figure~\ref{fig:NumOfFacetsAnal} presents the impact of varying the number of facet heads under top-$3$ routing. 
We observe that performance generally improves as the number of facet heads increases up to a moderate range, while further increasing the number of heads does not consistently provide additional gains.
For LaMP-3 in Figure~\ref{fig:NumOfFacetsAnal} (a), MoF generally achieves favorable performance when using 6--8 facet heads and performs competitively against the corresponding HYDRA variants across most settings. 
Similarly, for LaMP-4 in Figure~\ref{fig:NumOfFacetsAnal} (b), performance tends to improve up to around 6--7 facet heads before showing limited additional gains.
These observations suggest that increasing the number of facet heads alone does not necessarily improve personalization performance.
Additional results and analyses are presented in Appendix~\ref{app:SensitivityAnalysis}. 
% While additional facet heads increase model complexity and allow finer preference decomposition, excessive decomposition may reduce the effectiveness of shared preference modeling. 
% This indicates that a moderate number of facet heads provides a better balance between preference specialization and parameter sharing.

\begin{figure}[t]
  \centering
  \includegraphics[width=\columnwidth]{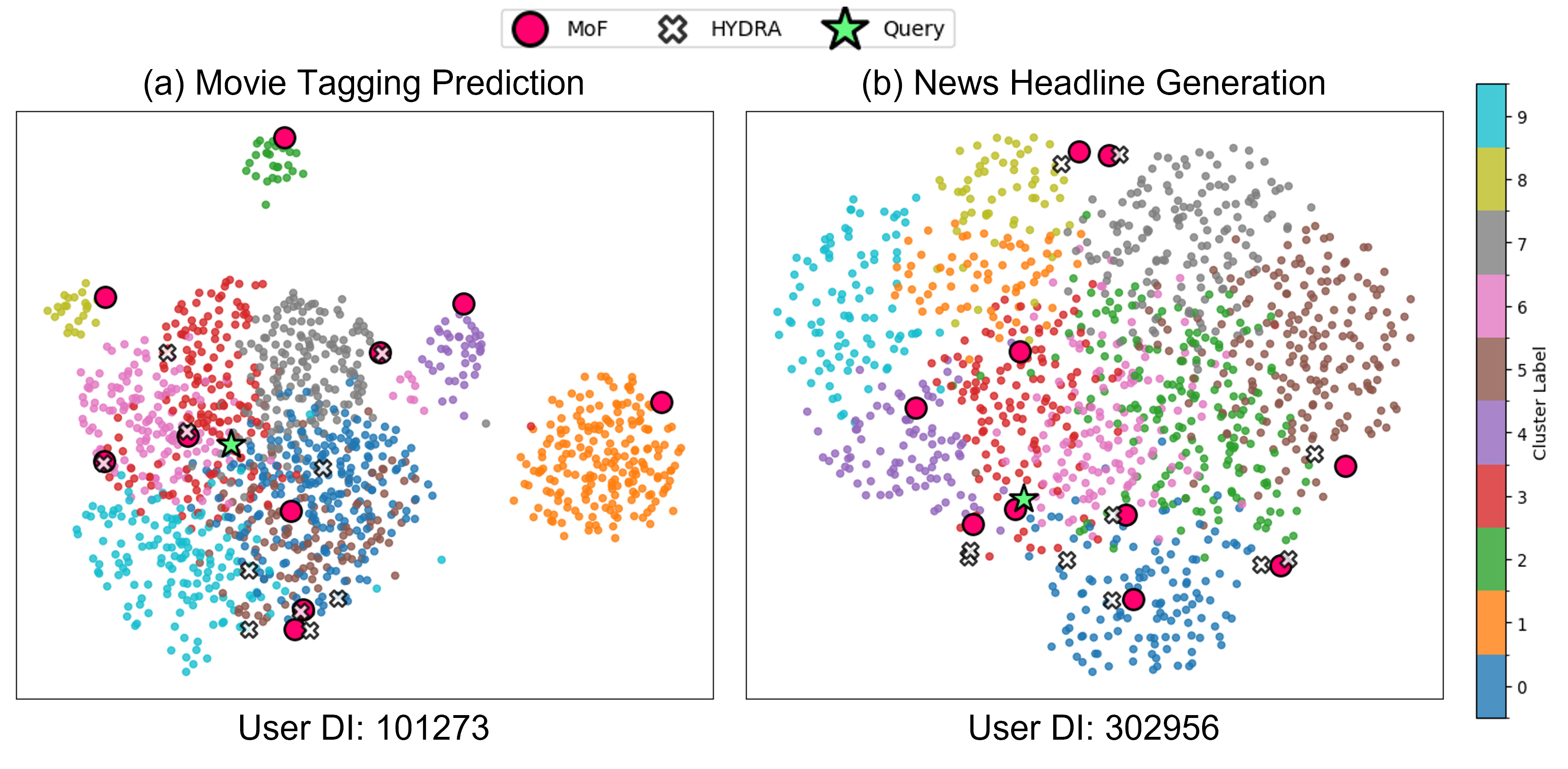}
  \caption{
  Visualization of CBS for two example users. Colors indicate clusters of user-history embeddings.
  Red circles and white crosses denote training samples selected using CBS in MoF and BM25 retrieval in HYDRA, respectively, and the green star denotes the query.
  }
  \label{fig:ClusterAnal}
\end{figure}
  
\subsection{Visualization}
\label{sec:Visualization}
Figure~\ref{fig:ClusterAnal} visualizes the history items selected for two example users, illustrating how CBS in MoF retrieves samples from multiple preference clusters compared with the standard retrieval used in HYDRA.
The visualization is obtained by projecting k-means-clustered history embeddings (K=10) using UMAP~\cite{mcinnes2018umap}.
Compared with HYDRA, which tends to retrieve histories concentrated within specific regions, MoF selects items across multiple clusters, exposing the model to more diverse preference signals.
We further observe that user histories often span multiple regions in the latent space (Figure~\ref{fig:ClusterAnal} (a)) rather than forming a single concentrated pattern.
Such heterogeneous preference distributions motivate CBS to provide richer signals for learning facet representations.
This observation is consistent with the ablation results, where removing CBS causes the largest performance degradation.

\subsection{Qualitative Analysis}
\label{sec:QualitativeAnalysis}
To investigate whether the learned router captures meaningful preference-related structures, we group users according to their routing-weight distributions and perform BERTopic~\cite{grootendorst2022bertopic} analysis on their historical interactions.
As illustrated in Figure~\ref{fig:RouterWeightAnal_2}, users exhibit distinct routing patterns rather than being assigned uniformly. 
We analyze the corresponding user groups in Table~\ref{tab:GroupTopicModelingLaMP2}, where users with similar routing patterns also show coherent narrative tendencies. 
For example, $G1$ is associated with threat- and conflict-centered narratives, whereas $G2$ focuses on action and fantasy themes, while $G3$ emphasizes character-driven and emotional narratives.
These findings provide qualitative evidence that routing weights reflect meaningful preference-related structures rather than arbitrary assignment patterns.

\section{Conclusion}
In this paper, we introduced MoF, a personalization framework for black-box LLMs based on shared preference facets and preference-aware routing. 
By replacing user-specific heads with a fixed number of shared facet heads, MoF achieves scalable personalization while maintaining constant personalization-parameter storage overhead as the number of users increases.
Experimental results further show that preference-aware routing supports effective personalization and generalization to users unseen during training without additional parameter updates.

\section*{Limitations}
While MoF achieves a substantially stronger average rank across tasks, it does not outperform HYDRA on every individual metric (e.g., BLEU on LaMP-4).
Thus, our results support strong overall personalization performance under a more scalable parameterization, rather than uniform superiority over HYDRA on every metric.

Although MoF generalizes to a substantially larger set of unseen users, our primary experiments train the model on a relatively limited number of users.
Therefore, the current results demonstrate inference-time scalability to larger user populations but do not fully characterize training behavior under substantially larger and more heterogeneous user populations.

MoF also employs a fixed number of shared facet heads for all users.
While this design provides parameter-efficient personalization, a fixed facet capacity may become restrictive as user preferences become increasingly diverse.
Future work could explore adaptive facet allocation or routing mechanisms that dynamically adjust model capacity while preserving parameter efficiency.

Finally, our evaluation is limited to the LaMP benchmark and does not include LongLaMP or a direct comparison with FERMI.
LongLaMP involves long-form personalization with task-specific profile structures that may require adapting the current MoF formulation, while FERMI performs iterative per-user prompt optimization under a different personalization paradigm.
Extending MoF to long-form settings and comparing it with such optimization-based approaches remain important directions for future work.

\section*{Ethical Considerations}
MoF performs personalization based on user interaction histories, which may contain sensitive information reflecting individual preferences and behavioral patterns.
Although MoF eliminates user-specific scoring heads and reduces the need for storing personalized parameters, the preference inference process may still capture sensitive behavioral patterns from historical interactions.
In addition, personalization systems may unintentionally reinforce existing user tendencies or reduce exposure to diverse content.
Practical deployments should therefore consider privacy-preserving mechanisms such as anonymization, secure handling of user histories, and user-controlled personalization settings.

We use publicly available artifacts according to their respective licenses and terms of use.
The LaMP benchmark code and data creation methods are licensed under CC BY-NC-SA 4.0, and we follow the corresponding licenses and usage conditions of the datasets included in the benchmark.
We use \texttt{bge-base-en-v1.5} (FlagEmbedding) under the MIT License.
We additionally use OpenAI models (e.g., \texttt{gpt-3.5-turbo-1106} and \texttt{gpt-4o-mini-2024-07-18}) according to OpenAI's API terms of use and service policies.

All artifacts were used in a manner consistent with their intended use and access conditions. 
Public benchmarks, pretrained models, and APIs were used only for research and evaluation purposes and in accordance with their specified usage policies.

\bibliography{references}

%%%%%%%%%%%%%%%%%%%%%%%%%%%%%%%%%%%%%%%%%%%%%%%%%%%%%%%%%%%%
\appendix

\section{Detailed Preliminary Framework}
\label{app:preliminary_details}

\subsection{Reranker}
The reranker $s_\phi^{\mathrm{R}}$ takes the current query $q_u$ and retrieved history candidates as input, and outputs a score indicating the usefulness of each history item for generating a user-preferred response.
For each user $u$, we first retrieve a set of $M$ history candidates $H'_u=\{h_{u,1}^{\prime},h_{u,2}^{\prime},\dots,h_{u,M}^{\prime}\}$ from the full user history $H_u$ based on their relevance to the target query $q_u$.
To construct the reranker training dataset $D^{\mathrm{R}}$, we randomly sample $M$ history items from $H_u$ and treat the query in each sampled history item as a target query for retrieval-based training instance construction.
We create a binary supervision label $r_{u,m}^{R} \in \{0,1\}$ indicating whether the history item $h_{u,m}^{\prime}$ helps generate a user-preferred response.
For classification tasks, a positive label is assigned when the model output matches the ground-truth label.
For generation tasks, a positive label is assigned when the evaluation score exceeds the average score across candidate responses.
The reranker is optimized using the following binary classification objective:
\begin{equation}
\mathcal{L}_{\mathrm{R}}(\phi)
=
\sum_{u \in \mathcal{U}}
\sum_{m=1}^{M}
\ell_{\mathrm{BCE}}
\Big(
s_\phi^{\mathrm{R}}(q_u,h_{u,m}^{\prime}),
r_{u,m}^{R}
\Big).
\end{equation}
Based on the reranker scores, the top-$k$ history items are selected to construct the personalized context set:
\begin{equation}
H_u^*
=
\arg\text{top-}k_{h_{u,m}^{\prime}\in H_u^{\prime}}
s_\phi^\mathrm{R}(q_u,h_{u,m}^{\prime})
\label{eq:reranker_inference}
\end{equation}

\subsection{Adapter}
Given the query $q_u$ and the personalized context set $H_u^*$, we sample a set of $B$ candidate responses $O_u=\{o_u^1,\dots,o_u^B\}$ from the black-box LLM.
The adapter $s_\theta^{\mathrm{A}}$ acts as a lightweight selection module that scores candidate responses without modifying the black-box LLM parameters, thereby aligning the output distribution with user preferences.
The adapter assigns a preference score to each candidate response in $O_u$ based on its alignment with user preferences.
To build the adapter training dataset $D^{\mathrm{A}}$, we generate candidate response sets $O_u$ for all history items in $H_u$.
We assign a binary supervision label $r_{u,b}^{A} \in \{0,1\}$ to each candidate response $o_u^b$, where $r_{u,b}^{A}=1$ indicates that the response aligns with user preferences.
Following the labeling strategy used for reranker training, positive labels are assigned based on exact matching for classification tasks and relative evaluation scores for generation tasks.
The adapter is optimized using the following binary classification objective:
\begin{equation}
\mathcal{L}_{\mathrm{A}}(\theta)
=
\sum_{u \in \mathcal{U}}
\sum_{b=1}^{B}
\ell_{\mathrm{BCE}}
\Big(
s_\theta^{\mathrm{A}}(q_u,o_u^b),
r_{u,b}^{A}
\Big).
\end{equation}
The final personalized response is selected as the candidate with the highest adapter score:
\begin{equation}
\hat{y}_u
=
\arg\max_b
s_\theta^{\mathrm{A}}(q_u,o_u^b).
\label{eq:adapter_inference}
\end{equation}

\subsection{Training \& Inference}
As illustrated in Figure~\ref{fig:ProcessOverview}, both the reranker $s_\phi^{\mathrm{R}}$ and the adapter $s_\theta^{\mathrm{A}}$ are trained in two stages: a train stage and a fit stage.
During the train stage, the base model and scoring heads are jointly trained using data collected from a large number of users.
This enables the model to capture shared preference signals across users while learning user-specific preference patterns through separate heads.
During the fit stage, only the heads are updated for unseen users while the shared base model remains frozen.
This stage adapts the personalization modules to the preference patterns of new users without modifying the shared parameters.

During inference, retrieved history items are first reranked to select the more useful historical items $H_u^*$ for the current query, as defined in Eq.~(\ref{eq:reranker_inference}).
Conditioned on $H_u^*$, the black-box LLM generates a set of candidate responses $O_u$.
The adapter then assigns preference scores to the candidate responses in $O_u$, and the response with the highest score is selected as the final personalized response $\hat{y}_u$, as defined in Eq.~(\ref{eq:adapter_inference}).

\section{Parameter Scalability Analysis}
\label{app:ParameterScalability}
We analyze how the number of trainable personalization parameters scales with the number of users.
Let $N$ denote the number of users, $F$ the number of facet heads.
For user-specific personalization approaches such as HYDRA, each user maintains an independent scoring head.
The trainable personalization parameters can therefore be expressed as
\begin{equation}
P_{\mathrm{HYDRA}}(N)
=
N P_{\mathrm{head}},
\end{equation}
where $P_{\mathrm{head}}$ denotes the parameter count of a single user-specific head.
Consequently, introducing one additional user increases the parameter count by
\begin{equation}
P_{\mathrm{HYDRA}}(N+1)
-
P_{\mathrm{HYDRA}}(N)
=
P_{\mathrm{head}}.
\end{equation}
Thus, the personalization parameter complexity grows linearly with the number of users:
\begin{equation}
P_{\mathrm{HYDRA}}(N)
=
O(N).
\end{equation}

In contrast, MoF employs a fixed set of shared facet heads coupled with a shared routing module.
The trainable personalization parameters become
\begin{equation}
P_{\mathrm{MoF}}
=
F P_{\psi}
+
P_{\mathrm{router}}
+
P_{\mathrm{SAE}},
\end{equation}
where $P_{\psi}$ denotes the parameter count of a single facet head, and $P_{\mathrm{router}}$ and $P_{\mathrm{SAE}}$ denote the parameters of the shared routing and sparse autoencoder modules, respectively.
Since all terms in $P_{\mathrm{MoF}}$ are independent of the number of users, introducing a new user does not increase the parameter count:
\begin{equation}
P_{\mathrm{MoF}}(N+1)
-
P_{\mathrm{MoF}}(N)
=
0.
\end{equation}
Therefore, MoF exhibits constant parameter complexity with respect to the number of users:
\begin{equation}
P_{\mathrm{MoF}}
=
O(1).
\end{equation}
This result indicates that personalization for unseen users can be achieved solely through history-derived routing weights without introducing additional user-specific parameters.
Consequently, MoF can scale to large and dynamically evolving user populations without increasing the number of trainable personalization parameters.

\section{Task Details}
\label{app:taskdetails}
We evaluate LLM personalization using the LaMP benchmark~\cite{salemi-etal-2024-lamp}.
The benchmark consists of seven personalized NLP tasks spanning both classification and generation settings.
We exclude LaMP-1 and LaMP-7 because these tasks involve task-specific input structures that are not directly compatible with the reranker \& adapter formulation adopted in our framework and prior black-box personalization methods. 
This exclusion reflects a limitation of the current reranker \& adapter instantiation rather than an inherent restriction of the MoF framework.
Since our goal is to evaluate personalized preference modeling under a unified black-box personalization setting, we focus on tasks that share comparable interaction structures and evaluation protocols. 
In addition, LaMP-6 is not publicly available and therefore cannot be included in our experiments. 
This evaluation setting is consistent with the evaluation setting used in prior black-box personalization work~\cite{zhuang2024hydra}.

\paragraph{LaMP-2: Movie Tagging Classification.}
LaMP-2 evaluates whether a model can predict a user's tagging behavior for movies.
Each user's history consists of a sequence of movie description--tag pairs, where the descriptions correspond to previously viewed movies and the tags represent the user's assigned labels.
The task requires the model to capture personalized preference patterns from the sequential interaction history and predict the appropriate tag for a target movie description.
Following prior work, we evaluate performance using Accuracy (Acc) and F1 score (F1).

\paragraph{LaMP-3: Product Rating Classification.}
LaMP-3 evaluates personalized rating prediction based on users' product reviews.
User history is represented as a sequence of review--rating pairs generated by the same user.
Given a target review, the model predicts the rating score that the user would assign.
Although LaMP-3 is formulated as an ordinal classification task in the original benchmark, following prior work we report MAE and RMSE as evaluation metrics.

\paragraph{LaMP-4: News Headline Generation.}
LaMP-4 evaluates personalized text generation by requiring the model to generate headlines for news articles.
User history consists of article--headline pairs written by the same journalist.
The task requires extracting the journalist's writing characteristics and stylistic patterns from historical interactions and generating a personalized headline for a target article.
We evaluate generation quality using ROUGE-1 (R-1), ROUGE-L (R-L), and BLEU.

\paragraph{LaMP-5: Scholarly Title Generation.}
LaMP-5 focuses on generating titles for academic papers based on their abstracts.
Similar to LaMP-4, each user's history consists of abstract--title pairs written by the same author.
The model is expected to infer the author's writing style and title construction patterns from historical data and generate a personalized title for a target abstract.
We evaluate performance using ROUGE-1 (R-1), ROUGE-L (R-L), and BLEU.

\begin{table}[t]
\caption{
Size of training data used in the train and fit stages of MoF across different LaMP datasets.
}
\centering
\small
\setlength{\tabcolsep}{4pt}
\renewcommand{\arraystretch}{1.0}

\resizebox{\columnwidth}{!}{
\begin{tabular}{l l l c c c c}
\toprule
\textbf{Method} &
\textbf{Module} &
\textbf{Stage} &
\textbf{\footnotesize LaMP-2} &
\textbf{\footnotesize LaMP-3} &
\textbf{\footnotesize LaMP-4} &
\textbf{\footnotesize LaMP-5} \\

\midrule

\multirow{4}{*}{MoF}
& \multirow{2}{*}{Reranker}
& Train & 10,228 & 10,852 & 10,816 & 10,998 \\
& & Fit & 4,468  & 4,926  & 4,841  & 4,999\\

\cmidrule(lr){2-7}

& \multirow{2}{*}{Adapter}
& Train & 167,979 & 140,195 & 289,368 & 71,226 \\
& & Fit & 45,401  & 76,336  & 128,871 & 39,294 \\

\bottomrule
\end{tabular}
}
\label{tab:mof_datasize}
\end{table}
\begin{table}[t]
\caption{
Training and fitting time of MoF across different LaMP datasets.
}
\centering
\small
\setlength{\tabcolsep}{4pt}
\renewcommand{\arraystretch}{1.0}

\resizebox{\columnwidth}{!}{
\begin{tabular}{l l l c c c c}
\toprule
\textbf{Method} &
\textbf{Module} &
\textbf{Stage} &
\textbf{\footnotesize LaMP-2} &
\textbf{\footnotesize LaMP-3} &
\textbf{\footnotesize LaMP-4} &
\textbf{\footnotesize LaMP-5} \\

\midrule

\multirow{4}{*}{MoF}
& \multirow{2}{*}{Reranker}
& Train & 00:04:40 & 00:09:37 & 00:03:36 & 00:09:52 \\
& & Fit & 00:00:56 & 00:01:14 & 00:00:58 & 00:01:19 \\

\cmidrule(lr){2-7}

& \multirow{2}{*}{Adapter}
& Train & 00:57:53 & 01:53:06 & 01:24:06 & 00:50:08 \\
& & Fit & 00:31:42 & 00:21:19 & 00:25:03 & 00:08:21 \\

\bottomrule
\end{tabular}
}
\label{tab:mof_runtime}
\end{table}

\begin{table*}[t]
\caption{
Performance on the LaMP benchmark using \texttt{gpt-4o-mini-2024-07-18} as the underlying proprietary black-box LLM.
Experimental settings follow Table~\ref{tab:MainResults}, while replacing the black-box LLM used in the main experiments.
Results indicate that MoF maintains strong overall performance, demonstrating robustness beyond a specific black-box model.
}
\centering
\small
\setlength{\tabcolsep}{4pt}
\renewcommand{\arraystretch}{1.1}
\resizebox{\textwidth}{!}{
\begin{tabular}{l cc cc ccc ccc c}
\toprule
\textbf{Dataset ($\rightarrow$)} 
  & \multicolumn{2}{c}{\textbf{LaMP-2}} 
  & \multicolumn{2}{c}{\textbf{LaMP-3}} 
  & \multicolumn{3}{c}{\textbf{LaMP-4}} 
  & \multicolumn{3}{c}{\textbf{LaMP-5}}
  & \multirow{2}{*}{\shortstack{\textbf{Avg.}\\\textbf{Rank}$\downarrow$}} \\
\cmidrule(lr){2-3} \cmidrule(lr){4-5} \cmidrule(lr){6-8} \cmidrule(lr){9-11}
\textbf{Method ($\downarrow$)} 
  & Acc~$\uparrow$ & F1~$\uparrow$ 
  & MAE~$\downarrow$ & RMSE~$\downarrow$ 
  & R-1~$\uparrow$ & R-L~$\uparrow$ & BLEU~$\uparrow$ 
  & R-1~$\uparrow$ & R-L~$\uparrow$ & BLEU~$\uparrow$ 
  & \\
\midrule

Zero-Shot
& 0.467 & 0.281 
& 0.320 & 0.639 
& 0.150 & 0.134 & 0.849
& 0.422 & 0.343 & 3.113
& 12.0 \\

\midrule

ICL ($k$=1)
& 0.513 & 0.344 
& 0.413 & 0.757 
& 0.150 & 0.129 & 0.908
& 0.434 & 0.364 & 4.525
& 12.0 \\

ICL ($k$=2)
& 0.493 & 0.365 
& 0.393 & 0.716 
& 0.152 & 0.135 & 1.250
& 0.433 & 0.364 & 4.043
& 11.2 \\

ICL ($k$=4)
& 0.520 & 0.388 
& 0.327 & 0.648 
& 0.156 & 0.140 & \textbf{1.681}
& 0.440 & 0.375 & 5.230
& 7.9 \\

\midrule

RAG ($k$=1)
& 0.520 & 0.349 
& 0.340 & 0.647 
& 0.157 & 0.141 & 0.806
& 0.432 & 0.359 & 4.204
& 10.4 \\

RAG ($k$=2)
& 0.527 & 0.361 
& 0.367 & 0.651 
& 0.165 & 0.145 & 1.425
& 0.461 & 0.385 & \underline{5.934}
& 7.6 \\

RAG ($k$=4)
& 0.573 & 0.405 
& 0.327 & 0.594 
& 0.163 & 0.147 & 1.304
& 0.466 & \textbf{0.410} & \textbf{6.078}
& 4.4 \\

\midrule

PAG ($k$=0)
& 0.507 & 0.370 
& \underline{0.287} & \underline{0.572} 
& 0.127 & 0.109 & 0.183
& 0.402 & 0.329 & 2.631
& 12.0 \\

PAG ($k$=1)
& 0.567 & 0.470 
& \textbf{0.280} & \textbf{0.566} 
& 0.142 & 0.123 & 0.354
& 0.399 & 0.340 & 2.890
& 10.2 \\

\midrule

CFRAG ($k$=4)
& \textbf{0.673} & \textbf{0.533} 
& 0.347 & 0.642 
& \underline{0.171} & \underline{0.148} & 1.441 
& 0.445 & 0.383 & 5.137 
& 5.3 \\

\midrule

HYDRA-Reranker
& 0.613 & 0.417 
& 0.347 & 0.635 
& 0.160 & 0.140 & 1.331
& 0.464 & 0.401 & 5.362
& 6.4 \\

HYDRA-Adapter
& 0.527 & 0.343 
& 0.307 & 0.610 
& 0.143 & 0.126 & 0.728
& 0.410 & 0.328 & 3.127
& 11.5 \\

HYDRA
& \underline{0.653} & 0.448 
& 0.347 & 0.632 
& 0.168 & 0.147 & 1.230
& \underline{0.468} & \underline{0.405} & 5.603
& 5.0 \\

\midrule

\rowcolor{gray!15}
MoF-Reranker (ours)
& 0.587 & 0.452 
& 0.320 & 0.611 
& 0.169 & \underline{0.148} & \underline{1.459}
& 0.462 & 0.403 & 5.252
& \underline{4.2} \\

\rowcolor{gray!15}
MoF-Adapter (ours)
& 0.507 & 0.336 
& 0.300 & 0.604 
& 0.152 & 0.132 & 1.116
& 0.419 & 0.345 & 2.985
& 10.5 \\

\rowcolor{gray!30}
MoF (ours)
& \underline{0.653} & \underline{0.504} 
& 0.327 & 0.627 
& \textbf{0.180} & \textbf{0.157} & 1.109
& \textbf{0.472} & \underline{0.405} & 5.643
& \textbf{3.6} \\

\bottomrule
\end{tabular}
}
\label{tab:GeneralizationToOtherModels}
\end{table*}

\section{Data Statistics}
Table~\ref{tab:mof_datasize} reports the number of training instances used in the train and fit stages for the reranker and adapter modules across different LaMP datasets.

\section{Computational Cost}
Table~\ref{tab:mof_runtime} reports the training and fitting runtime of the reranker and adapter modules across different LaMP datasets.

\section{Implementation Details}
\label{app:ImplementationDetails}
All experiments are conducted on a machine equipped with an AMD Ryzen Threadripper 1950X CPU and a single NVIDIA GeForce RTX 3090 GPU.
For black-box LLM generation, we set the temperature to $1.0$.
For training, we employ AdamW~\cite{loshchilov2018adamw} with a learning rate of $5.0\times10^{-5}$, a batch size of 64, and a weight decay of 0.01.
All models are trained for 2 epochs.
We set $\lambda = 0.5$, $\lambda_1 = 8.6\times10^{-4}$ and $\tau = 2$. 
For the reranker, we partition each user's history into 10 clusters and retrieve one item from each cluster using BM25, yielding \(M=10\) sampled candidates.
For the adapter, we set the number of candidate responses $B$ to 8.
Based on validation performance, we set the number of facet heads to $F=7$ and use top-3 sparse routing across all experiments.
To construct user embeddings, we use up to the most recent 128 history items for each user.
Additional sensitivity analyses are provided in Appendix~\ref{app:SensitivityAnalysis}.

\section{Generalization Across LLMs}
\label{app:GeneralizationToOtherModels}
Table~\ref{tab:GeneralizationToOtherModels} presents additional experimental results using \texttt{gpt-4o-mini-2024-07-18}~\cite{openai2024gpt4omini} as the underlying proprietary black-box LLM while keeping all other experimental settings identical to those in the main experiments. 
We observe that the overall trends remain largely consistent with those in Table~\ref{tab:MainResults}. 
Similar to the main results, prompt-based personalization methods do not consistently improve performance across tasks, whereas reranker \& adapter-based approaches generally provide stronger and more stable gains.
MoF maintains the strongest overall performance with the best average rank of 3.6, demonstrating strong performance across multiple tasks under a different black-box model setting. 
Specifically, MoF achieves the best performance on several evaluation metrics while remaining competitive on the remaining settings. 
For example, MoF improves F1 from 0.448 to 0.504 on LaMP-2 compared with HYDRA and also achieves stronger performance on LaMP-4, improving R-1 from 0.168 to 0.180 and R-L from 0.147 to 0.157. 
Although absolute performance differs from the main experimental setting, the relative advantage of MoF remains largely preserved.
These results suggest that the proposed framework can generalize across different proprietary black-box LLMs.

\section{Sensitivity Analysis}
\label{app:SensitivityAnalysis}
\subsection{Impact of Number of Retrieved Items}
\begin{figure}[t]
  \centering
  \includegraphics[width=\columnwidth]{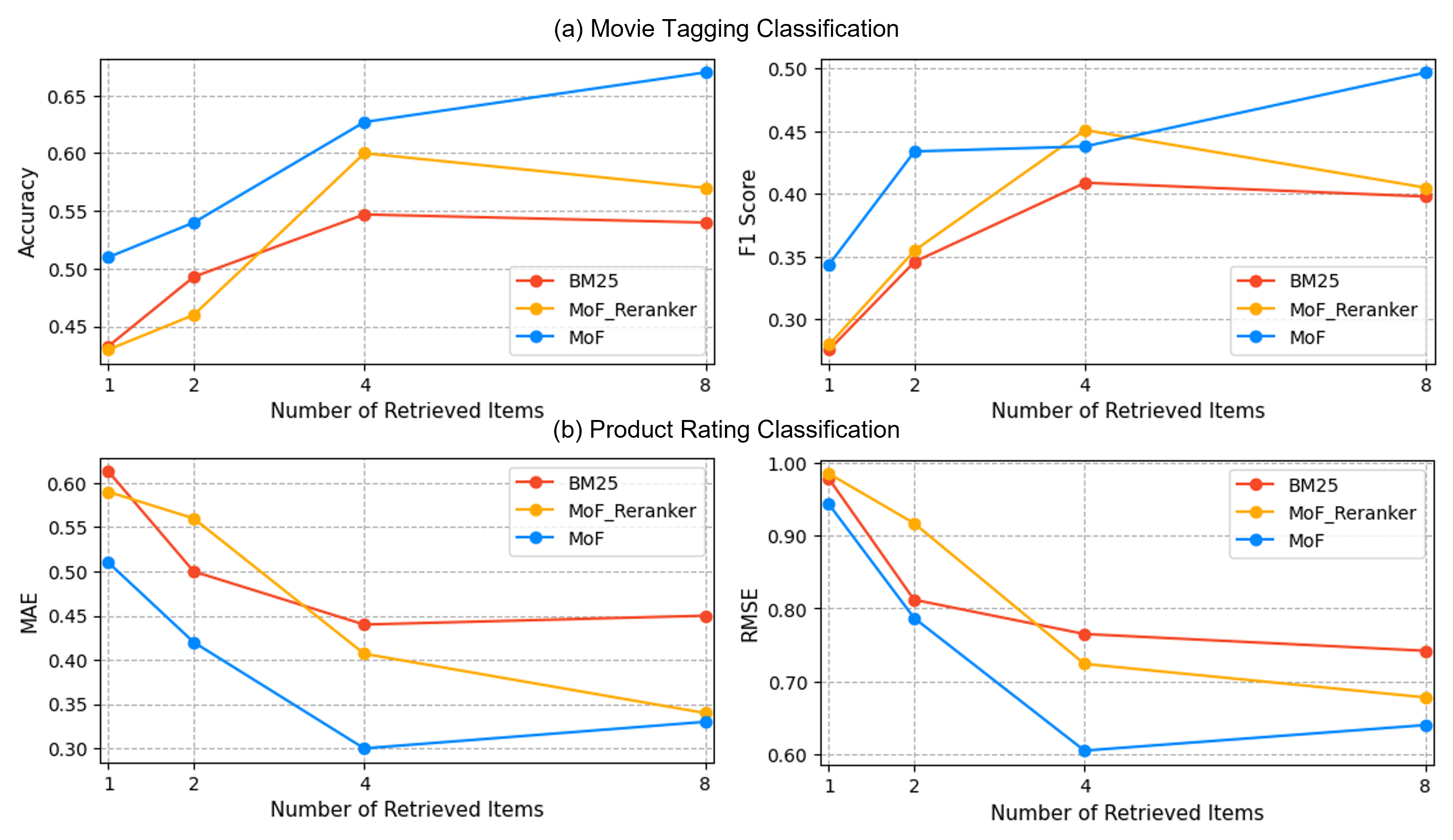}
  \caption{
  Impact of the number of retrieved user-history items on personalization performance.
  (a) shows results on LaMP-2 and (b) shows results on LaMP-3.
  MoF generally benefits from additional retrieved information up to moderate retrieval sizes and maintains strong performance across different retrieval sizes.
  }
  \label{fig:NumberofRetrievedItemAnal}
\end{figure}
Figure~\ref{fig:NumberofRetrievedItemAnal} presents the impact of varying the number of retrieved items on personalization performance.
We observe that increasing the number of retrieved items generally improves performance in both LaMP-2 and LaMP-3 up to a moderate retrieval size, while larger retrieval sizes may provide diminishing returns.
For LaMP-2 in Figure~\ref{fig:NumberofRetrievedItemAnal} (a), MoF shows consistent performance improvements as the number of retrieved items increases, whereas BM25 and MoF-Reranker exhibit weaker gains or performance saturation. 
This trend suggests that the proposed framework benefits more from additional retrieved information.
A similar trend can be observed for LaMP-3 in Figure~\ref{fig:NumberofRetrievedItemAnal} (b). 
Although performance slightly degrades at larger retrieval sizes, MoF maintains lower MAE and RMSE values than the baselines across different retrieval settings. 
These observations suggest that the proposed framework remains relatively robust under larger retrieval sizes while benefiting from richer user-history information.

\subsection{Impact of Top-k Routing}
\begin{figure}[t]
\centering
\includegraphics[width=\columnwidth]{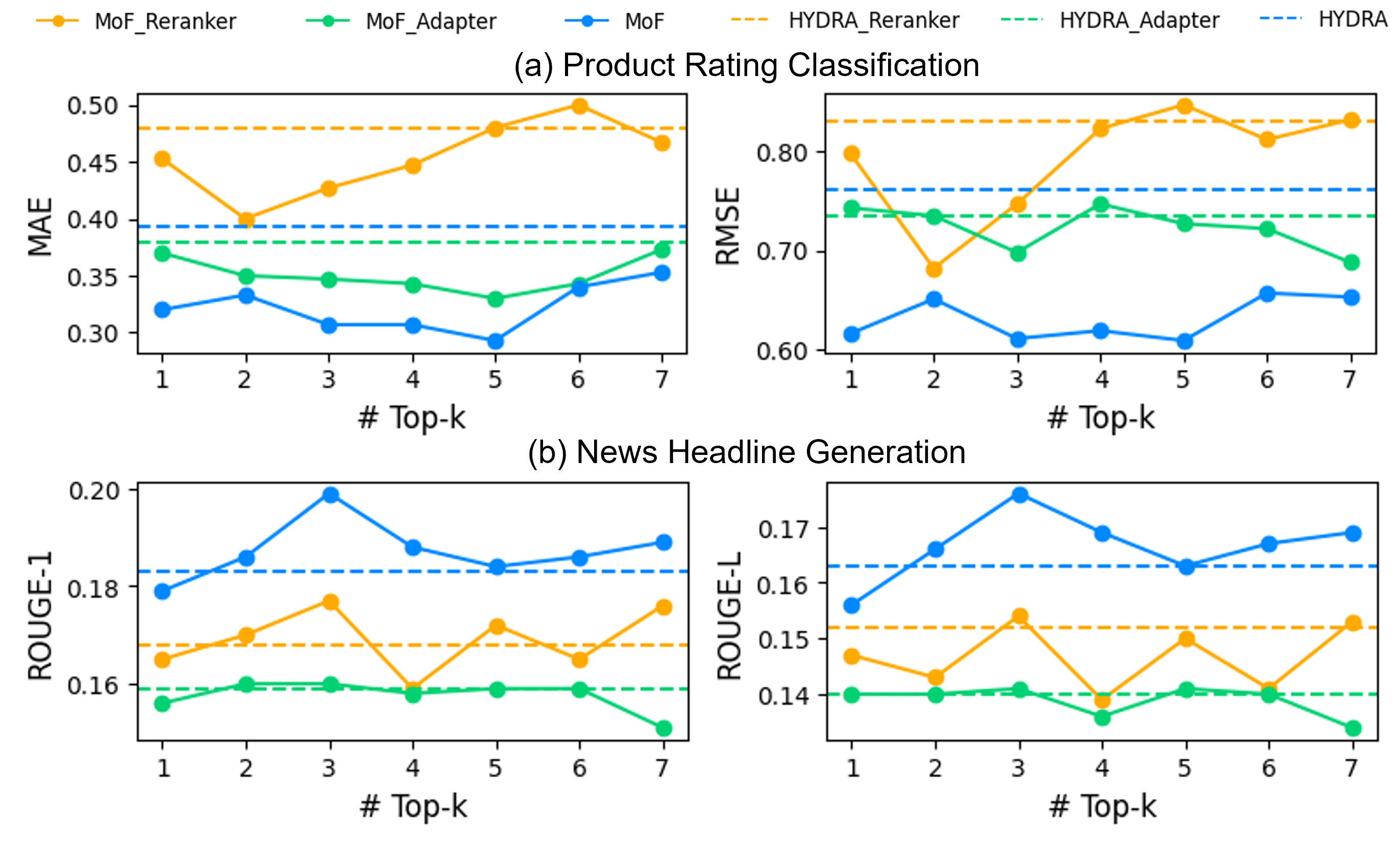}
\caption{
Impact of top-$k$ routing on personalization performance while fixing the total number of facet heads to 7.
(a) and (b) present results on LaMP-3 and LaMP-4, respectively.
Performance varies with the number of activated facet heads, indicating the importance of routing selection in preference aggregation.
} 
\label{fig:TopKAnal}
\end{figure}
Figure~\ref{fig:TopKAnal} presents the impact of varying the number of selected facet heads under top-$k$ routing while fixing the total number of facet heads to 7. 
We observe that performance generally improves up to a moderate routing size, while larger routing sizes provide limited additional benefits.
For LaMP-3 in Figure~\ref{fig:TopKAnal} (a), MoF achieves its best performance around k=2, obtaining lower MAE and RMSE values than the corresponding HYDRA variants across all settings. 
Similarly, for LaMP-4 in Figure~\ref{fig:TopKAnal} (b), performance reaches its highest values around k=3 for both R-1 and R-L metrics. 
In addition, MoF achieves stronger performance than HYDRA across most routing settings.
These observations suggest that selecting a moderate routing size provides a better balance between utilizing diverse preference signals and maintaining focused preference representations, which is consistent with prior MoE studies emphasizing sparse expert activation and effective expert specialization~\cite{zhou2022mixture}.
While activating additional facet heads may incorporate richer preference information, excessively large routing sizes may reduce the effectiveness of preference aggregation by introducing redundant or less relevant preference signals.

\begin{figure}[t]
  \centering
  \includegraphics[width=\columnwidth]{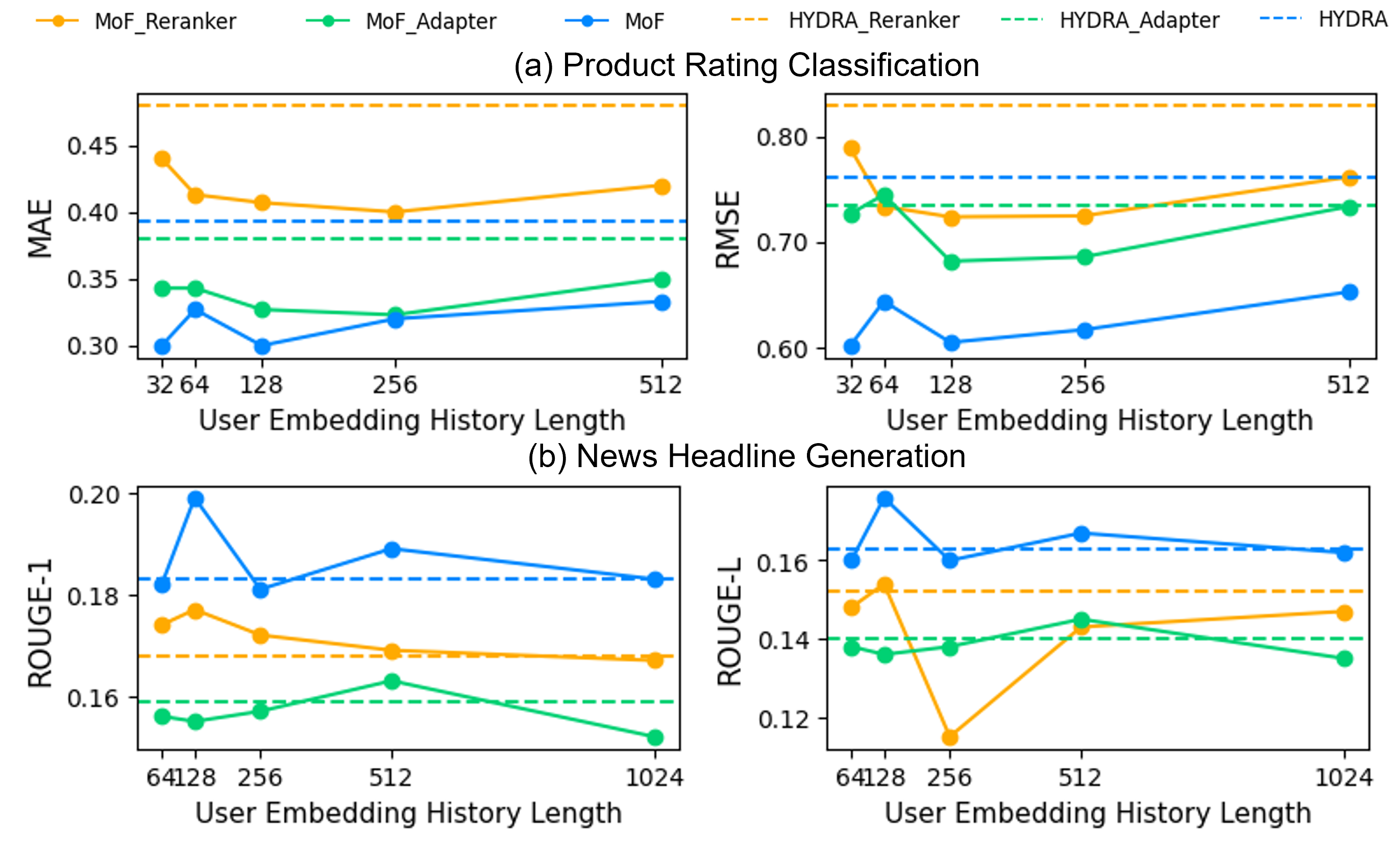}
  \caption{
  Impact of the number of recent user interactions used for user embedding construction on personalization performance.
  (a) and (b) present results on LaMP-3 and LaMP-4, respectively.
  Moderate history lengths generally provide stronger performance across tasks.
  }    
  \label{fig:UserHistoryLengthAnal}
\end{figure}

\subsection{Impact of User Embedding History Length}
Figure~\ref{fig:UserHistoryLengthAnal} presents the impact of varying the number of recent user interactions used for constructing user embeddings. 
We observe that increasing the history length does not consistently improve personalization performance.
For LaMP-3 in Figure~\ref{fig:UserHistoryLengthAnal} (a), MoF achieves its strongest performance at relatively short-to-moderate history lengths, with the best results observed around 128 recent interactions. 
Similarly, for LaMP-4 in Figure~\ref{fig:UserHistoryLengthAnal} (b), performance also peaks around moderate history lengths before showing limited additional gains. 
Across both tasks, MoF generally maintains stronger performance than the corresponding HYDRA variants.
These observations suggest that incorporating additional historical interactions does not necessarily provide proportional benefits. 
Prior work on sequential recommendation has shown that user representations benefit from focusing on behavior signals that are more relevant to current preferences rather than uniformly aggregating long interaction histories~\cite{kang2018sasrec}. 
Our results exhibit a similar trend, suggesting that moderate history lengths may provide a favorable balance between preserving relevant preference information and limiting the influence of older interactions.

\begin{figure*}[t]
  \centering
  \includegraphics[width=\textwidth]{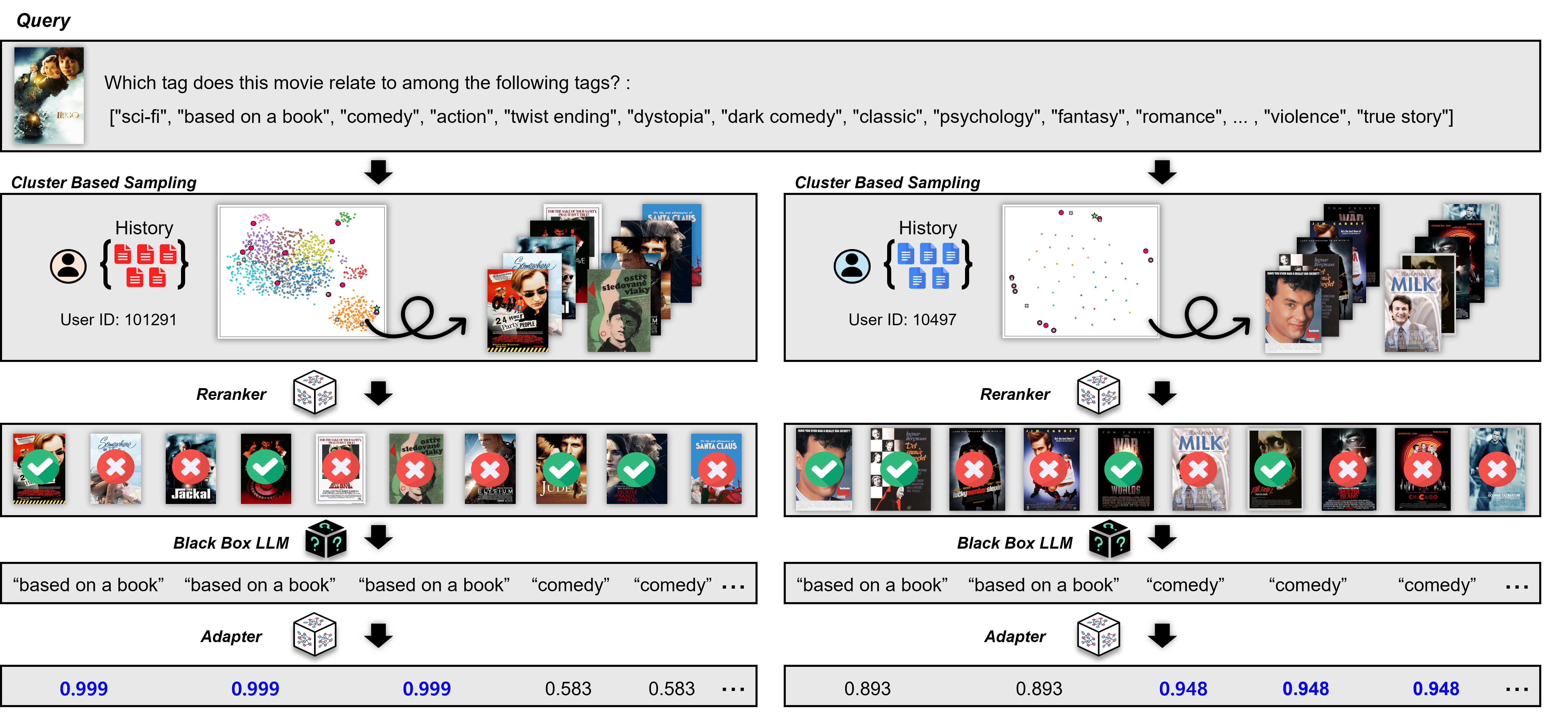}
  \caption{
  Case study on LaMP-2 illustrating how MoF personalizes prediction behavior for the same query.
  MoF selects different user-history examples and assigns distinct preference alignment scores according to each user's tagging patterns, resulting in user-specific evidence selection and predictions.
  }
  \label{fig:CaseStudy}
\end{figure*} 

\begin{figure*}[t]
  \centering
  \includegraphics[width=\textwidth]{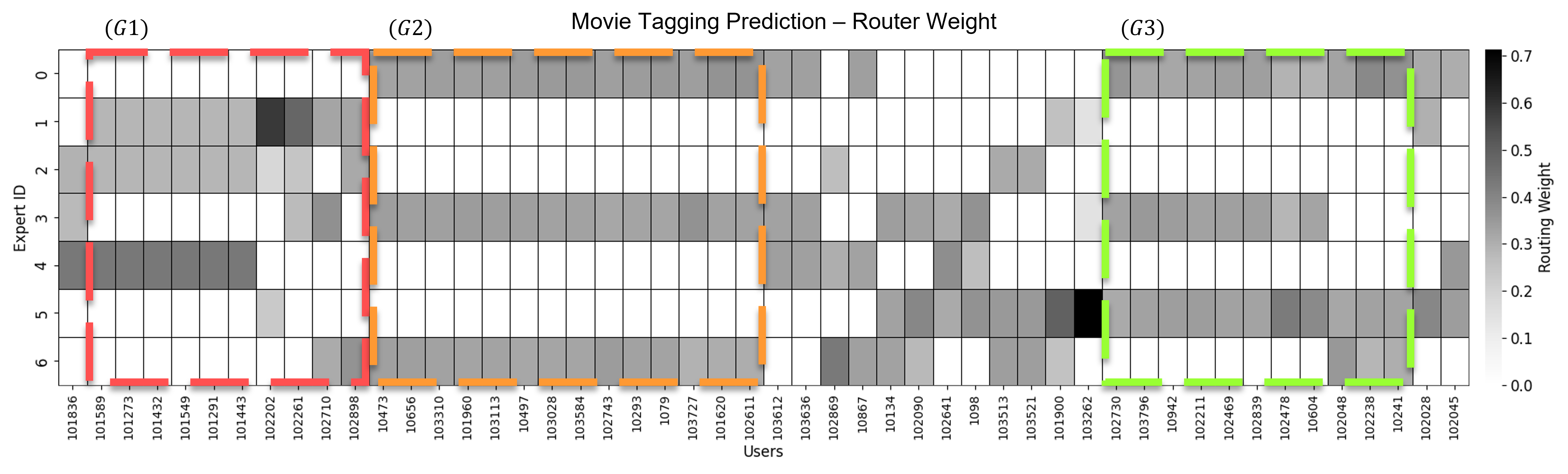}
  \caption{
    Heatmap of routing weight distributions across users. Users with similar routing patterns are grouped together.
    }
  \label{fig:RouterWeightAnal_2}
\end{figure*}

\section{Case Study}
\label{app:CaseStudy}
Figure~\ref{fig:CaseStudy} presents qualitative examples from LaMP-2 to illustrate how MoF performs preference-aware personalization under the same input query. 
In this task, the model is asked to predict the movie tag for the same target movie from a candidate tag set, while each user provides a different interaction history. 
Although the target query is identical, MoF retrieves different user-history samples and assigns different preference alignment scores depending on the user's past tagging behavior.

For User \texttt{101291}, MoF selects historical movies that are mostly associated with the tag ``based on a book.'' 
The selected examples are distributed across multiple regions in the history embedding space, indicating that MoF does not simply rely on local semantic similarity, but instead gathers preference-relevant evidence from diverse historical clusters. 
As a result, the model assigns the highest scores to the ``based on a book'' candidates, suggesting that the user's previous tagging patterns provide strong evidence for this preference.

In contrast, for User \texttt{10497}, MoF retrieves a different set of historical examples under the same target query. 
The selected histories include multiple examples associated with ``comedy,'' and the resulting preference scores are highest for the ``comedy'' candidates. 
This shows that MoF can adapt its prediction behavior to a different user's preference pattern even when the input movie is unchanged. 
The comparison demonstrates that MoF personalizes the decision process by conditioning on user-specific historical preferences rather than producing a single globally preferred prediction.

Overall, this case study highlights two important properties of MoF. 
First, the cluster-based history selection encourages the model to use diverse evidence from the user's interaction history, which helps capture heterogeneous preference signals. 
Second, the preference-aware routing mechanism enables different users to emphasize different latent facets, leading to user-specific score distributions and final predictions. 
These qualitative results support the main experimental findings that MoF can effectively model personalized preferences while avoiding user-specific scoring heads.

\section{Prompt Templates}
\begin{figure*}[t]
\caption{
Prompt template used for Movie Tagging Classification (LaMP-2).
The prompt incorporates retrieved user history examples together with task instructions and the target movie description.
}
\centering
\includegraphics[width=\textwidth]{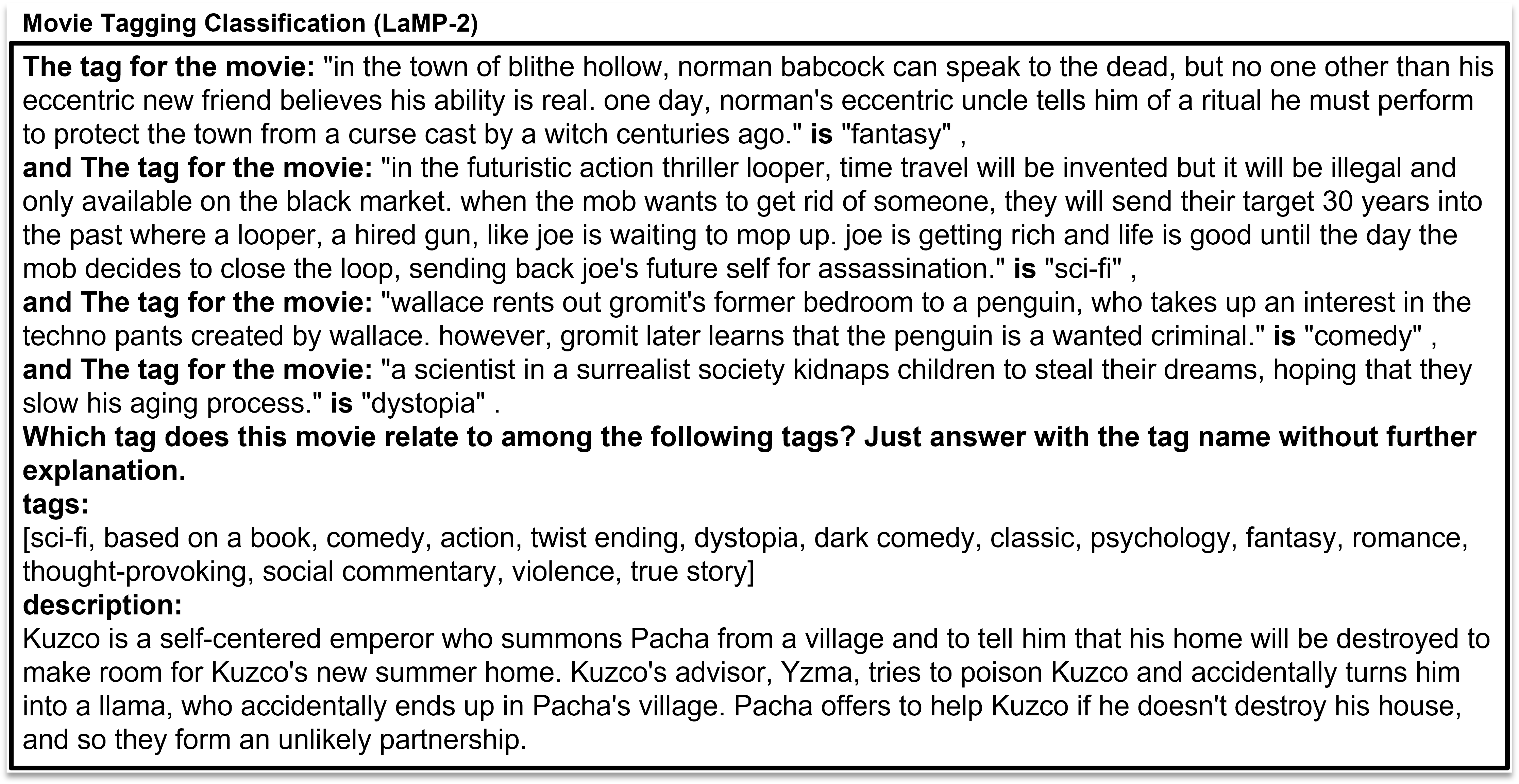}
\label{fig:PromptLaMP2}
\end{figure*}

\begin{figure*}[t]
\caption{
Prompt template used for Product Rating Classification (LaMP-3).
The prompt incorporates retrieved user history examples together with task instructions and the target review.
}
\centering
\includegraphics[width=\textwidth]{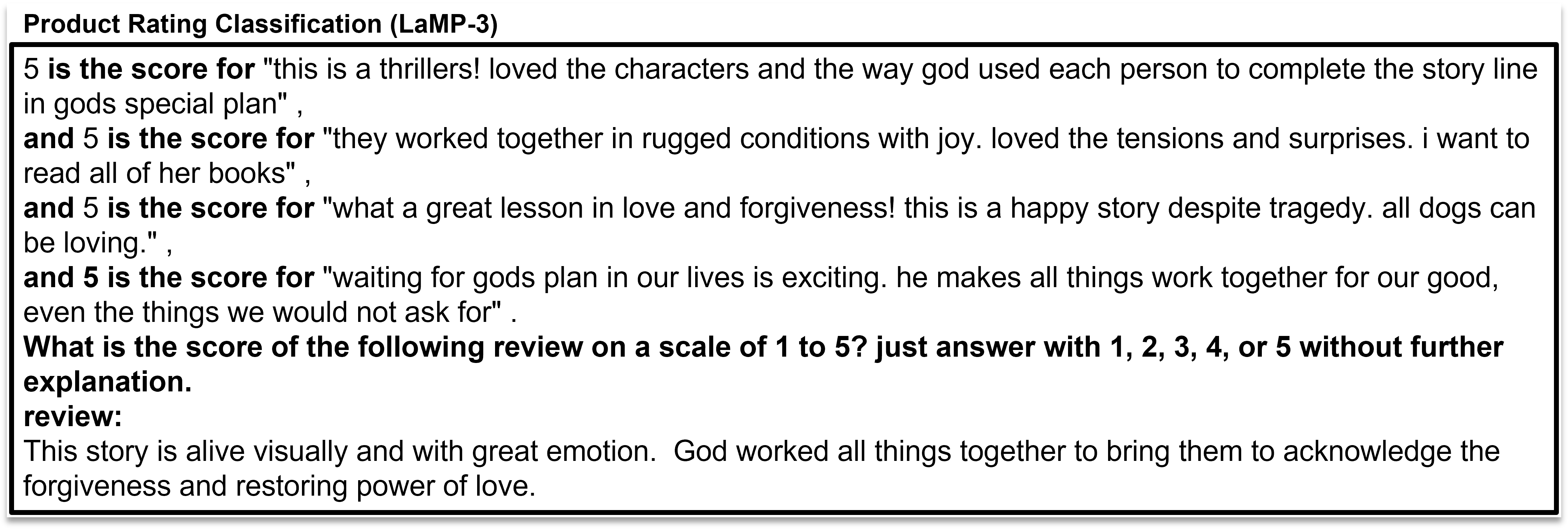}
\label{fig:PromptLaMP3}
\end{figure*}

\begin{figure*}[t]
\caption{
Prompt template used for News Headline Generation (LaMP-4).
The prompt incorporates retrieved user history examples together with task instructions and the target news article.
}
\centering
\includegraphics[width=\textwidth]{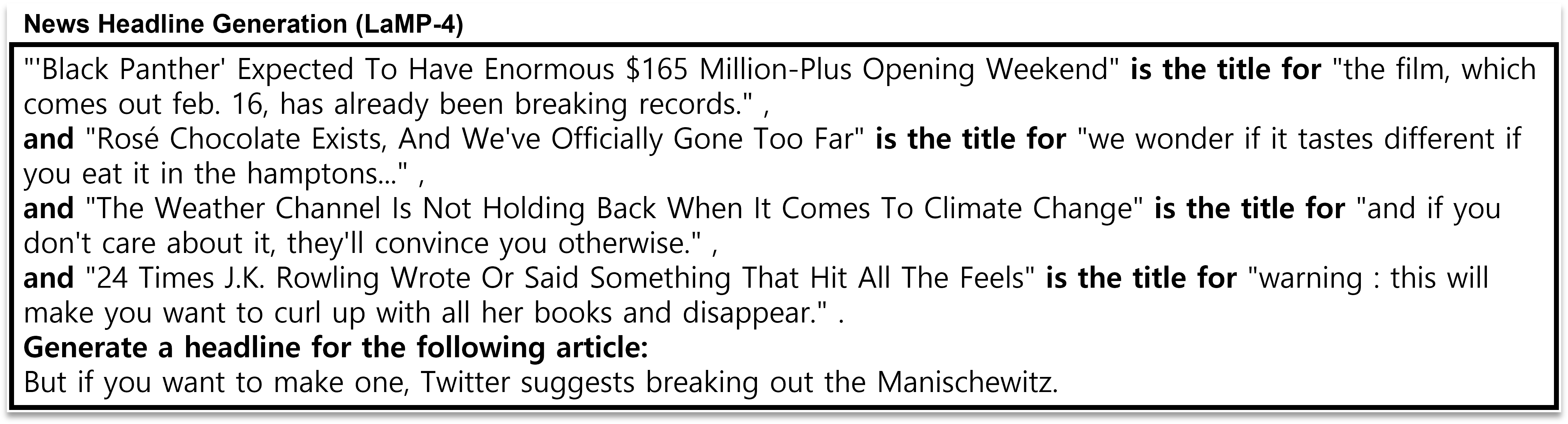}
\label{fig:PromptLaMP4}
\end{figure*}

\begin{figure*}[t]
\caption{
Prompt template used for Scholarly Title Generation (LaMP-5).
The prompt incorporates retrieved user history examples together with task instructions and the target paper abstract.
}
\centering
\includegraphics[width=\textwidth]{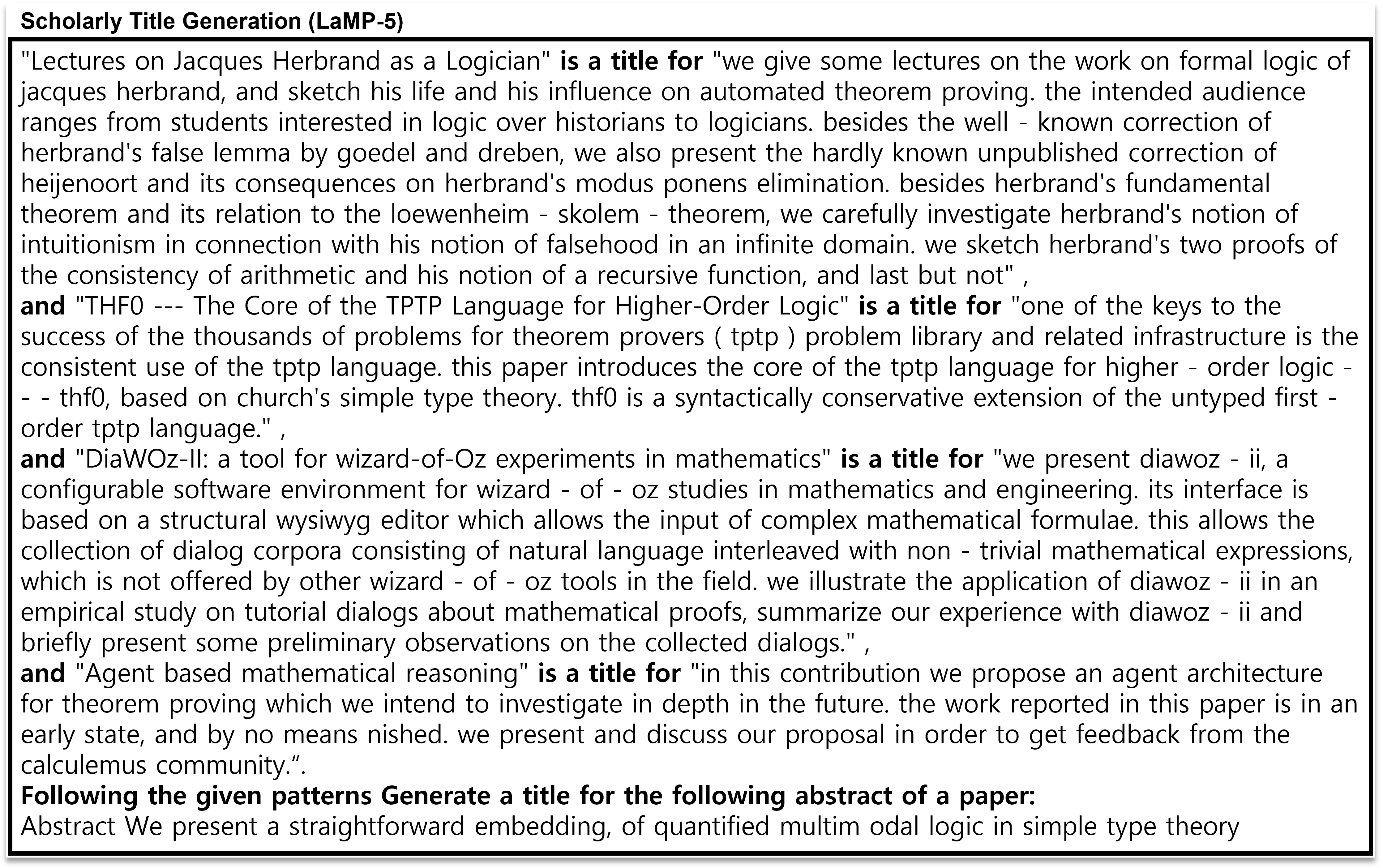}
\label{fig:PromptLaMP5}
\end{figure*}

Following prior work~\cite{zhuang2024hydra,salemi-etal-2024-lamp}, we employ prompt templates throughout the framework, including black-box LLM generation and the construction of training instances for both the reranker and adapter modules.
The prompts combine task-specific instructions with personalized contextual information derived from user histories.
The complete prompt templates used in our experiments are illustrated in Figures~\ref{fig:PromptLaMP2}, \ref{fig:PromptLaMP3}, \ref{fig:PromptLaMP4}, and \ref{fig:PromptLaMP5}.

\end{document}